\documentclass[11pt]{article}

\usepackage[T1]{fontenc}
\usepackage{lmodern}

\usepackage[margin=1in]{geometry}
\usepackage[english]{babel}
\usepackage{appendix}

\usepackage{amsmath,amssymb,amsthm,mathtools,mathrsfs,bbm}
\usepackage{relsize}

\usepackage{float}
\usepackage{graphicx}
\usepackage{subcaption}
\usepackage{booktabs}
\usepackage{placeins}
\usepackage{tikz}
\usetikzlibrary{plotmarks}
\usepackage{pgfplots}
\usepgfplotslibrary{groupplots}
\pgfplotsset{compat=1.18}
\usepgfplotslibrary{fillbetween}

\usepackage{caption}
\usepackage{array}
\usepackage{tabularx}

\definecolor{empiricalred}{HTML}{C92640}
\definecolor{theoryblue}{HTML}{1960A5}

\usepackage{placeins}
\usepackage{algorithm}
\usepackage{algorithmic}

\usepackage{enumitem}
\usepackage{xcolor}
\usepackage[]{natbib}
\usepackage{fancyhdr}
\usepackage{titlesec}
\usepackage{pifont}
\usepackage{hyperref}

\hypersetup{
    colorlinks=true,
    linkcolor=black,
    citecolor=black,
    urlcolor=blue
}
\allowdisplaybreaks

\newcommand{\eps}{\varepsilon}
\renewcommand{\epsilon}{\eps}

\newtheorem{thm}{Theorem}

\newtheorem{lem}{Lemma}
\newtheorem{cor}{Corollary}

\newtheorem{rem}{Remark}

\titleformat{\section}{\large\bfseries}{\thesection}{1em}{}
\title{Nearly Minimax-Optimal Sample Complexity of Natural Policy Gradient with a Matching Information-Theoretic Lower Bound}

\author{%
Asha Barua\textsuperscript{1}, 
Sajad Khodadadian\textsuperscript{1}
}

\date{}

\begin{document}

\maketitle
\thispagestyle{plain}

\begingroup
\renewcommand{\thefootnote}{\fnsymbol{footnote}}
\endgroup
\footnotetext[1]{Grado Department of Industrial and Systems Engineering,
Virginia Polytechnic Institute and State University,
Blacksburg, Virginia 24061, USA.
E-mail: \texttt{\{ashabarua,sajadk\}@vt.edu}.}

\setcounter{footnote}{1}

\begin{abstract}
A fundamental question in reinforcement learning is how many sampled transitions are needed to learn a near-optimal policy, and whether direct policy optimization can attain this statistical limit. To answer this question, we consider natural policy gradient (NPG), a
widely used policy optimization method. We study tabular
finite-horizon Markov decision processes (MDPs) with known rewards and unknown horizon-dependent transition kernels, accessed through a generative model. Our sample-based NPG algorithms update the policy by
backward sweeps and do not estimate transition probabilities. Under
three different batch-size and step size schedules, we establish
last-iterate guarantees with sample complexity
$\widetilde O(SAH^4/\varepsilon^2)$, where $S$, $A$, and $H$ denote
the number of states, the number of actions, and the horizon length, respectively. In this setting, horizon-dependent transition kernels contribute a factor of $H$ to our transition-query bound. Our information-theoretic lower bound of $\Omega(SAH^4/\varepsilon^2)$
shows that the resulting $H^4$ dependence is unavoidable in the worst case, even for agents that select their queries adaptively. To the best of our knowledge, these are the first last-iterate guarantees for NPG whose sample complexity matches an
information-theoretic lower bound up to logarithmic factors. Consequently, our results demonstrate that direct policy-space optimization through NPG can achieve the fundamental statistical limits of reinforcement learning while guaranteeing the performance of
the final policy. Furthermore, for known kernels, we prove that exact NPG with a constant step size $\eta=1/H$ requires $\Omega(H^2/\varepsilon)$ iterations in the worst case,
matching the known upper bound for that step size.
\end{abstract}

\section{Introduction}
\label{sec:introduction}
Reinforcement learning (RL) provides a framework for sequential decision making under uncertainty and has achieved empirical success in game playing, robotic control,
autonomous navigation, and large-scale decision systems
\citep{singh2022reinforcement, silver2016mastering,mnih2015human, 8103164}. A Markov decision process (MDP) provides the mathematical foundation for RL, which describes an agent's interaction
with an uncertain environment. The objective is to learn a policy that maximizes the expected cumulative reward \citep{puterman2014markov,sutton2018reinforcement}.

Policy optimization methods pursue this objective by searching directly
over a class of policies. Policy gradient (PG) methods carry out this
search through gradient ascent on the expected return of a parameterized policy, an approach that accommodates both structured and high-dimensional policy classes \citep{williams1992simple,sutton1999policy}. Among these
methods, natural policy gradient (NPG), introduced by \citet{kakade2001natural}, accounts for the geometry of the policy space through the Fisher information metric. This geometric viewpoint has influenced practical methods such as trust region policy optimization (TRPO) and proximal policy optimization (PPO) \citep{schulman2015trust, schulman2017proximal}. In the tabular softmax setting, the multiplicative NPG update is
equivalent to a Kullback-Leibler (KL)-regularized policy mirror
descent (PMD) update over the simplex of action distributions at
each state \citep{shani2020adaptive,lan2023policy}. Consequently, convergence guarantees for this PMD update also hold
for the multiplicative NPG update. This
equivalence connects the analysis of NPG to the broader theory of
policy optimization.

NPG and its variants have achieved empirical success in complex control
tasks, including high-dimensional robotic manipulation
\citep{peters2008natural,rajeswaran2018learning}. These applications
raise a basic question about sample efficiency. How many transitions
does NPG need to learn a near-optimal policy? A substantial body of
work has established convergence and sample-complexity guarantees for
policy optimization methods, including NPG, in discounted
infinite-horizon MDPs
\citep{agarwal2021theory,mei2020global,cen2022fast,xiao2022convergence,
lan2023policy,johnson2023optimal}. In the discounted infinite-horizon setting,
a single transition kernel governs the dynamics, a stationary optimal
policy exists, and the Bellman optimality operator is a contraction.
For tasks with a fixed number of decisions, however, a finite-horizon MDP
provides a natural formulation \citep{azar2017minimax,jin2018q,
dann2017unifying,jin2020provably}. In a finite-horizon MDP, optimal decision rules, value functions, and
transition kernels can vary across horizons, which introduces a distinct
theoretical challenge for policy optimization.  For this class of MDPs, \citet{barua2026finitetime}
established finite-time convergence guarantees for NPG with known action
values. We study the corresponding sample-based problem under generative
access, with known rewards and unknown transition probabilities. Our NPG method maintains a tabular policy and its action-value estimates in $\mathcal{O}(SAH)$ space without estimating transition probabilities. Our goal is to determine how many transition queries NPG needs for its final policy to achieve an expected optimality gap under a fixed initial distribution.

Prior work on finite-horizon MDPs provides several sample-complexity
benchmarks. \citet{sidford2018near} establish a high-probability
guarantee for variance-reduced value iteration when the transition kernel is shared across horizons. For horizon-dependent kernels, \citet{li2024breaking} establish sharp sample-complexity bounds for
model-based planning. \citet{li2022minimax} analyze Q-FTRL for Markov games and, in the single-agent setting, give a high-probability guarantee for a weighted average of policy iterates. These results leave open whether policy-based methods such as NPG can attain the minimax sample complexity, up to logarithmic factors, with a guarantee for its final policy.

Our main contributions are as follows:
\begin{enumerate}
    \item \textbf{Iteration lower bound for exact NPG:}
    We establish a worst-case iteration lower bound of
    $\Omega(H^2/\varepsilon)$ for exact NPG with constant
    step size. This matches the dependence on
    $H$ and $\varepsilon$ in the corresponding upper bound of
    \citet{barua2026finitetime}.

    \item \textbf{Last-iterate guarantees for sample-based NPG:}
    We show that under three step size schedules (geometrically increasing, constant and linearly
    decreasing) $\widetilde{O}(SAH^4/\varepsilon^2)$ transition queries suffice
    for the final policy of sample-based NPG to achieve an expected optimality gap of at
    most $\varepsilon$.

    \item \textbf{Information-theoretic benchmark:}
    For every $S,A,H\ge2$, we establish that any algorithm requires $\Omega(SAH^4/\varepsilon^2)$ transition queries under
    the same expected-gap criterion. The bound
    allows adaptive query selection and imposes no memory restriction. Consequently, the proposed NPG procedures attain last-iterate guarantees with near-minimax-optimal sample complexity up to
    logarithmic factors.
\end{enumerate}

\section{Related Literature}
\label{sec:related-work}

Introduced more than two decades ago by \citet{kakade2001natural}, NPG and its equivalent form PMD have received substantial theoretical attention in recent years. For $\gamma$-discounted infinite-horizon MDPs, \citet{johnson2023optimal} obtained a dimension-independent linear convergence rate of $\gamma$ for exact PMD with adaptive step sizes and proved a matching worst-case lower bound for the class of PMD
methods they analyzed. For finite-horizon MDPs, \citet{barua2026finitetime} established finite-time guarantees for exact NPG.
\begingroup

\newcommand{\npgrate}[2]{%
  \ensuremath{\widetilde{O}\!\left(\frac{#1}{#2}\right)}}
\newcommand{\npgbound}[2]{%
  \ensuremath{O\!\left(\frac{#1}{#2}\right)}}
\newcommand{\npgrange}{\ensuremath{(0,(1-\gamma)^{-1})}}
\newcommand{\npgsource}[1]{\par #1\par}
\newcommand{\npgheavyrule}{\noalign{\hrule height 0.8pt}}
\newcommand{\npgtablehead}{%
  \npgheavyrule
  Algorithm
  & Sample complexity
  & $\varepsilon$-range
  & Guarantee
  & \shortstack{Last\\iterate} \\
  \npgheavyrule}

\renewcommand{\tabularxcolumn}[1]{m{#1}}

\FloatBarrier
\begin{table*}[!ht]
\centering
\normalsize
\setlength{\tabcolsep}{3pt}
\setlength{\arrayrulewidth}{0.35pt}
\setlength{\extrarowheight}{1.5pt}
\renewcommand{\arraystretch}{1.05}

\begin{tabularx}{\linewidth}{@{}
  >{\centering\arraybackslash}m{.285\linewidth}|
  >{\centering\arraybackslash}m{.235\linewidth}|
  >{\centering\arraybackslash}m{.135\linewidth}|
  >{\centering\arraybackslash}X|
  c@{}}
\npgtablehead

Inexact PMD
\npgsource{\citet[Thm.~16]{xiao2022convergence}}
  & \npgrate{SA}{(1-\gamma)^8\varepsilon^2}
  & \npgrange
  & High-probability bound 
  & $\checkmark$ \\
\hline

Inexact PMD
\npgsource{\citet[Thm.~5.1]{johnson2023optimal}}
  & \npgrate{SA}{(1-\gamma)^8\varepsilon^2}
  & \npgrange
  & High-probability bound 
  & $\checkmark$ \\
\hline

Stochastic PMD
\npgsource{\citet[Prop.~2]{lan2023policy}}
  & \npgrate{SA}{(1-\gamma)^5\varepsilon^2}
  & \npgrange
  & Expected gap 
  & $\times$ \\
\hline

Stochastic block PMD
\npgsource{\citet[Thm.~4.4]{lan2023block}}
  & \npgrate{SA}{(1-\gamma)^5\varepsilon^2}
  & \npgrange
  & Expected gap 
  & $\times$ \\
\hline

Lookahead PMD
\npgsource{\citet[Thm.~5.4]{protopapas2024policy}}
  & \npgrate{SA}{(1-\gamma)^7\varepsilon^2}
  & \npgrange
  & High-probability bound 
  & $\checkmark$ \\
\hline

TD-PMD
\npgsource{\citet[Thm.~4.3]{liu2025temporal}}
  & \npgrate{SA}{(1-\gamma)^7\varepsilon^2}
  & \npgrange
  & High-probability bound 
  & $\checkmark$ \\
\hline

Sample-based NPG and Q-NPG
\npgsource{\citet{agarwal2021theory}}
  & \npgbound{1}{(1-\gamma)^{11}\varepsilon^6}
  & $\varepsilon>0$
  & Expected minimum gap 
  & $\times$ \\
\hline

Log-linear NPG and Q-NPG
\npgsource{\citet{yuan2023linear}}
  & NPG: \npgrate{1}{(1-\gamma)^5\varepsilon^2}
    \par
    Q-NPG: \npgrate{1}{(1-\gamma)^7\varepsilon^2}
  & $\varepsilon>0$
  & Expected gap 
  & $\checkmark$ \\
\hline

Accelerated NPG
\npgsource{\citet[Thm.~1]{mondal2024improved}}
  & \npgbound{1}{(1-\gamma)^6\varepsilon^2}
  & Sufficiently small $\varepsilon>0$
  & Average of the expected gaps 
  & $\times$ \\
\hline

Stochastic dynamic PG
\npgsource{\citet{klein2025structure}}
  & \npgrate{1}{(1-\gamma)^{17} \delta^{10}\varepsilon^8}
  & \npgrange
  & High-probability bound 
  & $\checkmark$ \\
\npgheavyrule
\end{tabularx}

\caption[Model-free policy optimization in discounted infinite-horizon MDPs]{%
Sample-complexity bounds for model-free policy optimization in
$\gamma$-discounted infinite-horizon MDPs under generative access.
Each rollout step counts as one transition query. The gap is the
optimality gap. For guarantees with an additive approximation error,
$\varepsilon$ denotes the allowable gap above that error.
A check mark indicates that the guarantee applies to the last iterate.}
\label{tab:npg-discounted-policy-optimization}
\end{table*}

When action values are estimated from sampled transitions, the relevant
measure of efficiency is the total number of transition queries needed
to learn a near-optimal policy. Beyond NPG and PMD, our comparison
covers sample-complexity guarantees for other model-free policy
optimization methods that can be implemented under generative access.
Table~\ref{tab:npg-discounted-policy-optimization} summarizes these
guarantees for discounted infinite-horizon MDPs. Table~\ref{tab:npg-finite-policy-optimization} compares our
last-iterate NPG bounds with prior guarantees for finite-horizon MDPs
with horizon-dependent transition dynamics. Throughout both tables, $\widetilde{O}$ suppresses logarithmic factors.

In Table~\ref{tab:npg-discounted-policy-optimization}, the sampling assumptions
vary across methods. Some allow the algorithm to query any chosen
state-action pair, while others use rollouts initialized from a
prescribed distribution over states or state-action pairs.
With per-step rewards in $[0,1]$, the discounted optimality gap
is at most $(1-\gamma)^{-1}$, so the nontrivial accuracy range is
$0<\varepsilon<(1-\gamma)^{-1}$.

\FloatBarrier
\begin{table*}[!ht]
\centering
\normalsize
\setlength{\tabcolsep}{3pt}
\setlength{\arrayrulewidth}{0.35pt}
\setlength{\extrarowheight}{1.5pt}
\renewcommand{\arraystretch}{1.05}

\begin{tabularx}{\linewidth}{@{}
  >{\centering\arraybackslash}m{.285\linewidth}|
  >{\centering\arraybackslash}m{.235\linewidth}|
  >{\centering\arraybackslash}m{.135\linewidth}|
  >{\centering\arraybackslash}X|
  c@{}}
\npgtablehead

Simultaneous stochastic softmax PG
\npgsource{\citet[Thm.~4.1]{klein2023beyond}}
  & \npgrate{S^8H^{41}}{\delta^{10}\varepsilon^8}
  & $(0,H)$
  & High-probability bound 
  & $\checkmark$ \\
\hline

Dynamic stochastic softmax PG
\npgsource{\citet[Thm.~4.2]{klein2023beyond}}
  & \npgbound{A^{18}H^{28}}{\delta^{10}\varepsilon^8}
  & $(0,H)$
  & High-probability bound 
  & $\checkmark$ \\
\hline

NPG with a geometrically increasing step size
\npgsource{\textbf{This work}}
  & \npgrate{SAH^4}{\varepsilon^2}
  & $(0,H]$
  & Expected gap 
  & $\checkmark$ \\
\hline

NPG with a constant step size
\npgsource{\textbf{This work}}
  & \npgrate{SAH^4}{\varepsilon^2}
  & $(0,1]$
  & Expected gap 
  & $\checkmark$ \\
\hline

NPG with a linearly decreasing step size
\npgsource{\textbf{This work}}
  & \npgrate{SAH^4}{\varepsilon^2}
  & $(0,1]$
  & Expected gap 
  & $\checkmark$ \\
\npgheavyrule
\end{tabularx}

\caption[Model-free policy optimization in finite-horizon MDPs]{%
Sample-complexity bounds for model-free policy optimization in
finite-horizon tabular MDPs with generative access, horizon-dependent
transition kernels, and per-step rewards in $[0,1]$.
For \citet{klein2023beyond}, we compute the total number of transition
queries by summing, over all updates, the number of rollouts at each
update times their length. Their guarantees hold with probability at
least $1-\delta$. Our NPG methods return a last iterate $\pi_T$ that
is $\varepsilon$-optimal in expectation under the fixed initial
state distribution $\rho$.}
\label{tab:npg-finite-policy-optimization}
\end{table*}
\endgroup

In Table~\ref{tab:npg-finite-policy-optimization}, $H\ge2$ denotes
the horizon length. The
stochastic softmax PG methods of \citet{klein2023beyond} start
rollouts from prescribed state distributions. Our NPG methods sample
successor states by querying the generative model at each state-action
pair and each nonterminal horizon. In our result for NPG with a geometrically
increasing step size, both the batch size $m_t=2^t$ and the step size
$\eta_t=m_t/H$ double at each iteration. The accuracy range for this
schedule includes $\varepsilon=H$ because per-horizon rewards lie in
$[0,1]$, so every policy has an optimality gap of at most $H$. The
bounds for the constant and linearly decreasing step sizes hold for
$A\ge2$ and $0<\varepsilon\le1$.

For a fixed state-action pair $(s,a)$, the distribution of the
successor state, $P^h(\cdot\mid s,a)$, may vary with $h$. At each
nonterminal horizon $h\in[H-1]$, our NPG methods use
$\widetilde{O}(SAH^3/\varepsilon^2)$ transition queries across all
state-action pairs and iterations. At the terminal horizon $H$, the
action values equal the known terminal rewards and can therefore be
evaluated directly. Summing the per-horizon query bound over the
$H-1$ nonterminal horizons gives the total transition-query bound
$\widetilde{O}(SAH^4/\varepsilon^2)$. In a stationary discounted MDP,
a single transition kernel governs all transitions. Hence, the sample complexity of NPG for finite horizon MDPs has an extra horizon dependence $H$ compared to the $(1-\gamma)^{-1}$ dependency of NPG in infinite-horizon MDPs.

\section{Problem Formulation} \label{sec:problem-formulation}

\subsection{Finite-horizon Markov decision processes}
\label{subsec:finite-horizon-mdp}
We consider a finite-horizon Markov decision process (MDP) represented by $M(\mathcal{S},\mathcal{A},H,\mathcal{P},\mathcal{R})$, where
$\mathcal{S}$ and $\mathcal{A}$ are finite, nonempty state and action sets,
respectively, and $H \geq 2$ is the horizon length. We write $S=|\mathcal{S}|$,
$A=|\mathcal{A}|$, $[H]=\{1,\ldots,H\}$, and for a finite set $\mathcal{X}$,
$\Delta(\mathcal{X})$ is the probability simplex over $\mathcal{X}$. The transition kernels are $ \mathcal{P}
  =\left\{
    P^h:\mathcal{S}\times\mathcal{A}\to\Delta(\mathcal{S})
    \mid h\in[H-1]
  \right\}$, where $P^h(s'\mid s,a)$ is the probability of transitioning to state $s'$ after taking action $a$ in state $s$ at horizon $h$. The reward functions are $\mathcal{R}
  =\left\{
    R^h:\mathcal{S}\times\mathcal{A}\to[0,1]
    \mid h\in[H]
  \right\}$.

A nonstationary Markov policy is $\pi=(\pi^1,\ldots,\pi^H)$, where
$\pi^h(\cdot\mid s)\in\Delta(\mathcal A)$ is the action
distribution at state $s$ and horizon $h$.  Given $h\in[H]$ and $S^h=s$,
a policy $\pi$ generates the trajectory
$(S^h,A^h,\ldots,S^H,A^H)$ with
$A^i\sim\pi^i(\cdot\mid S^i)$ for $i\in\{h,\ldots,H\}$ and
$S^{i+1}\sim P^i(\cdot\mid S^i,A^i)$ for
$i\in\{h,\ldots,H-1\}$.
For $h\in[H]$ and $s\in\mathcal{S}$, the value function is defined as 
\[
V^{\pi,h}(s)
:=
\mathbb{E}\!\left[
\sum_{i=h}^{H}R^i(S^i,A^i)
\,\middle|\,
S^h=s
\right],
\]
where the expectation is over the trajectory generated by policy $\pi$. The corresponding action-value function is
\begin{align*}
Q^{\pi,h}(s,a)
&:=
R^h(s,a)
+
\sum_{s'\in\mathcal{S}}
P^h(s'\mid s,a)\,V^{\pi,h+1}(s'),
\qquad h\in[H-1],
\end{align*}
with terminal condition $
Q^{\pi,H}(s,a)
:=
R^H(s,a).$ For every $h\in[H]$ and $s\in\mathcal{S}$, the value and action-value functions satisfy
$V^{\pi,h}(s)=
\sum_{a\in\mathcal{A}}\pi^h(a\mid s)\,Q^{\pi,h}(s,a)$.
Since the rewards belong to $[0,1]$, we have 
\[
0\le V^{\pi,h}(s)\le H-h+1,
\qquad
0\le Q^{\pi,h}(s,a)\le H-h+1.
\]
For every $\pi,\ h\in[H]\ \text{and}\ s\in\mathcal S$, the goal is to find an optimal policy $\pi^\star$ that satisfies
\[
V^{\pi^\star,h}(s)\ge V^{\pi,h}(s).
\]
A superscript $\star$ denotes quantities associated with an optimal
policy $\pi^\star$. In particular, we write 
$V^{\star,h}:=V^{\pi^\star,h}$ and
$Q^{\star,h}:=Q^{\pi^\star,h}$. For $1 \le h\le i\le H$ and
$s,\widetilde s\in\mathcal{S}$, the state visitation probability under policy $\pi$ is
   \[
   d_s^{h\to i,\pi}(\widetilde s)
   :=\mathbb P\!\left(S^i=\widetilde s\,\middle|\,S^h=s\right),
   \]
   where the trajectory is initialized at $s$ at horizon $h$ and follows policy $\pi$. In particular,
   $d_s^{h\to h,\pi}(\widetilde s)=\mathbf 1\{\widetilde s=s\}$. For a distribution $\rho\in\Delta(\mathcal{S})$, we write
$V^{\pi,h}(\rho):=\sum_{s\in\mathcal{S}}\rho(s)V^{\pi,h}(s)$. 

Throughout the paper, $h,i,j$ index horizons,
$t$ indexes iterations, $i',j'$ index decision states
in the constructed family, and $k$ indexes enumerated actions.

\subsection{NPG Update Rule}
We use the multiplicative form of tabular softmax natural policy gradient (NPG) \citep{kakade2001natural}. The corresponding formulation in the discounted setting is given by \citet[Lemma~15]{agarwal2021theory}. NPG starts from the uniform policy
$\pi_0^h(a\mid s)=1/A$ for every
$(h,s,a)\in[H]\times\mathcal S\times\mathcal A$. For $t \geq 0$, let $\pi_t$ denote the policy after $t$ iterations. Given a step size $\eta_t>0$,
the exact NPG updates each decision rule according to
\begin{equation}
\label{eq:npg-update}
  \pi_{t+1}^h(a\mid s)
  =\frac{
    \pi_t^h(a\mid s)
    \exp\left(\eta_t Q^{\pi_t,h}(s,a)\right)
  }{Z_t^h(s)},
\end{equation}
where the normalization constant is
$Z_t^h(s)
  :=\sum_{a'\in\mathcal{A}}
    \pi_t^h(a'\mid s)
    \exp\left(\eta_t Q^{\pi_t,h}(s,a')\right)$. For a constant step size, we write $\eta_t \equiv \eta > 0$. At each iteration, exact NPG evaluates all horizons under the current policy $\pi_t$ and then updates their decision rules jointly. For every finite step size, the update depends smoothly on the
action values and preserves the support of the current policy. Uniform initialization therefore ensures full support at every finite iteration. In the tabular setting, as $\eta\to\infty$, the NPG update approaches
the greedy action-selection step of classical policy iteration \citep{puterman1994markov}, applied to the action values (whether exact or estimated) used in the update.

\subsection{A constructed family of finite-horizon MDPs}
\label{subsec:common-mdp-family}
Our family adapts the construction used by
\citet[Theorem~2.5.2]{kakade2003sample}. Both constructions have an
absorbing state with a positive reward, along with a set of decision states with zero reward. In our construction, every action has a positive probability
of entering the reward state on the next transition. The transition
parameters may vary across horizons, and the favorable action (which attains the maximum expected cumulative reward)
may vary across both states and horizons. Thus, the same decision
state can have different favorable actions at different horizons.

Suppose that $\mathcal S=\{s_{i'}:i'\in[K]\}\cup\{g\}$, $\mathcal A=\{a_k:k\in[A]\}$ where $s_1,\ldots,s_K$ are decision states and $g$ is an absorbing
reward state. Thus, $|\mathcal S|=K+1=S$.
The state and action spaces are shared across all horizons. Let $\rho\in\Delta(\mathcal S)$ be an arbitrary initial distribution
satisfying $\rho(g)=0$. The same distribution is used for all instances in the family, and no uniformity assumption is imposed on its probabilities over the decision states.
For every $h\in[H]$, $i'\in[K]$, and $a\in\mathcal A$, we define $R^h(s_{i'},a)=0,\;R^h(g,a)=1.$ For $p\in(0,1)$, define $B_p\in\Delta(\mathcal S)$ by
\begin{equation}
\label{eq:mixing-distribution}
B_p(g)=p,
\qquad
B_p(s_{j'})=\frac{1-p}{K},
\quad j'\in[K].
\end{equation}
For each $h\in[H-1]$, choose parameters $p^h$ and $\alpha^h$
satisfying
\begin{equation}
\label{eq:transition-parameter-restrictions}
0<\alpha^h<p^h,
\qquad
p^h+\alpha^h\le\frac1{2H}.
\end{equation}
These conditions admit parameter choices for every $H\ge2$. Let
\[
\Theta
:=
\left\{
\theta=(\theta_{i'}^h)_{h\in[H-1],\,i'\in[K]}
:
\theta_{i'}^h\in\{0,1,\ldots,A\}
\right\}.
\]
A label $\theta_{i'}^h=k\in[A]$ identifies $a_k$ as the favorable
action at state $s_{i'}$ and horizon $h$. The label
$\theta_{i'}^h=0$ means that all actions have the same transition
distribution at that state and horizon. Now, we fix an admissible parameter sequence
$(p^h,\alpha^h)_{h\in[H-1]}$. For each $\theta\in\Theta$, define
\begin{equation}
\label{eq:labelled-transition-probability}
p_{\theta,i',a_k}^h
:=
p^h-\alpha^h
+
2\alpha^h\mathbf1\{\theta_{i'}^h=k\},
\end{equation}
and set
\begin{equation}
\label{eq:family-transition-kernel}
P_\theta^h(\cdot\mid s_{i'},a_k)
:=
B_{p_{\theta,i',a_k}^h},
\qquad
h\in[H-1],\ i'\in[K],\ k\in[A].
\end{equation}
At the reward state, for every $h\in[H-1]$, $a\in\mathcal A$, and $j'\in[K]$, we set $P_\theta^h(g\mid g,a)=1,\;P_\theta^h(s_{j'}\mid g,a)=0.$ Consequently, once the process reaches $g$, it remains there and receives reward one at each remaining horizon. The parameter restrictions ensure that
$p_{\theta,i',a_k}^h\in(0,1)$, so the construction defines valid transition kernels. When a favorable action is present, it enters $g$ with probability
$p^h+\alpha^h$, whereas every other action enters $g$ with
probability $p^h-\alpha^h$. When the label is zero, every action
enters $g$ with probability $p^h-\alpha^h$. From any decision
state, conditional on not entering $g$, the next decision state
is uniformly distributed over $\{s_1,\ldots,s_K\}$, regardless of
the current state and action. 

For a fixed admissible parameter sequence $(p^h,\alpha^h)_{h\in[H-1]}$
and label array $\theta\in\Theta$, write $M_\theta:=M(\mathcal S,\mathcal A,H,\mathcal P_\theta,\mathcal R),$ where $\mathcal P_\theta=\{P_\theta^h:h\in[H-1]\}$ and
$\mathcal R=\{R^h:h\in[H]\}$. Let $\mathcal M_{K,A,H}$ denote the
family obtained by varying the admissible parameter sequence and the
label array. Each result below specifies its choice of parameters
and any additional restrictions on the labels.

Since $\rho(g)=0$ and the probability of entering $g$ from a decision
state is at most $1/(2H)$ per transition, the probability of remaining
in the set of decision states through horizon $h$ is at least
$1-(h-1)/(2H)\ge 1/2$ for every $h\in[H]$ and every policy.
At each decision state $s_{i'}$ and nonterminal horizon $h$ with
$\theta_{i'}^h\ne0$, the favorable action has a strictly higher
action value than every other action under any continuation policy.
By \eqref{eq:family-action-gap}, this gap is of order
$\alpha^h(H-h)$, where $H-h$ is the number of reward periods
remaining after the current decision. Therefore, all actions have equal action
values at state--horizon pairs with label zero, at the absorbing
reward state, and at the terminal horizon.

\section{Iteration Lower Bound for Exact NPG}
\label{sec:MDP}

We first consider the setting in which the transition kernels and
reward functions are known, so that exact NPG can evaluate the
action-value functions exactly. The following theorem gives a
worst-case lower bound on the number of iterations required to
achieve a prescribed last-iterate optimality gap over the family
constructed in Section~\ref{subsec:common-mdp-family}.

\begin{thm}[Iteration lower bound for exact NPG]
\label{thm:exact-npg-lower-bound}
Fix integers $S,A,H\ge 2$ and $T\ge 1$, and let $K:=S-1$.
Consider exact NPG with uniform initialization and constant step
size $\eta>0$. There exists an instance $M\in\mathcal M_{K,A,H}$
such that, for every initial distribution $\rho\in\Delta(\mathcal S)$
satisfying $\rho(g)=0$, the policy $\pi_T$ after $T$ iterations
satisfies
\[
V^{\star,1}(\rho)-V^{\pi_T,1}(\rho)
\ge
\frac{1}{128}
\min\!\left\{H,\frac{H\log A}{\eta T}\right\}.
\]
Consequently, when $\eta=1/H$ and $0<\varepsilon<H/128$,
guaranteeing a last-iterate optimality gap of at most $\varepsilon$
on every instance in $\mathcal M_{K,A,H}$ requires
\[
T\ge \frac{H^2\log A}{128\varepsilon}.
\]
\end{thm}
The proof in Appendix~\ref{app:exact-npg-lower-bound} constructs
an instance for each iteration budget $T$ and step size $\eta$.
On that instance, the lower bound holds simultaneously for all
initial distributions supported on the decision states.

\begin{rem}(Relation to the existing upper bound)
\label{rem:exact-npg-upper-lower}
Under uniform policy initialization and constant step size $\eta=1/H$,
\citet[Theorem~1]{barua2026finitetime} give an upper bound of
$H^2(1+\log A)/T$ on the last-iterate optimality gap.
Since $1+\log A\le (1+1/\log 2)\log A$ for every $A\ge2$,
this upper bound and Theorem~\ref{thm:exact-npg-lower-bound}
give matching iteration bounds up to universal constants.
Thus, for every $K\ge1$, $A\ge2$, $H\ge2$, and every initial
distribution $\rho$ supported on the decision states, the worst-case
iteration complexity of exact NPG with this initialization and step
size over $\mathcal M_{K,A,H}$ is $\Theta\!\left(\frac{H^2\log A}{\varepsilon}\right)$ where $0<\varepsilon<\frac{H}{128}$.
\end{rem}

\section{Sample-based Setting}
\label{sec:sample-based}

We consider unknown horizon-dependent transition kernels and known reward functions. We introduce the sampling framework and NPG update, then analyze three choices of batch and step size schedules. The section concludes with an information-theoretic lower bound.

Generative access is a standard benchmark for studying the sample
complexity of reinforcement learning
\citep{kearns1999finite,kakade2003sample,azar2013minimax,li2024breaking}.
The agent can query any triple
$(h,s,a)\in[H-1]\times\mathcal S\times\mathcal A$.
Conditional on the query and all preceding information, the oracle
returns a fresh successor with distribution
$P^h(\cdot\mid s,a)$. Queries can depend on preceding observations
and the agent's internal randomness. Each successor observation
counts as one transition query, and the agent uses at most $N$
queries almost surely. No query is required at horizon $H$,
since $Q^{\pi,H}=R^H$ is known.

We adopt the storage-based interpretation of model-free algorithms
in \citet[Definition~1]{jin2018q}. The proposed NPG algorithms use $\mathcal{O}(SAH)$ memory
throughout execution, including the returned policy, without estimating transition probabilities. Stored estimates
are updated and reused without retaining all previous iterates
or the full sample history. 

Each iteration performs a backward sweep of NPG updates through
horizons $H,H-1,\ldots,1$. Let $\mathcal F_t$ denote the information
available before iteration $t$, including all previous oracle
observations and algorithmic randomness revealed up to that point.
The current policy $\pi_t$, the positive integer batch size $m_t$,
and the positive step size $\eta_t$ are $\mathcal F_t$-measurable. For a continuation value $v:\mathcal S\to\mathbb R$, define
\begin{equation}
\label{eq:expected-continuation-value}
(P^h v)(s,a)
:=
\sum_{s'\in\mathcal S}P^h(s'\mid s,a)v(s'),
\qquad h\in[H-1].
\end{equation}
At horizon $h<H$ of iteration $t$, draw $m_t$ fresh successors
$Y_{t,1}^{h,s,a},\ldots,Y_{t,m_t}^{h,s,a}$ at each
$(s,a)\in\mathcal S\times\mathcal A$, and define
\begin{equation}
\label{eq:sampled-continuation-value}
(\bar P_t^h v)(s,a)
:=
\frac{1}{m_t}\sum_{\ell=1}^{m_t}
v\!\left(Y_{t,\ell}^{h,s,a}\right).
\end{equation}
Here, $\ell$ indexes samples within the batch, and $\bar P_t^h$
denotes the corresponding sampling operator. This average is
computed by accumulating sampled successor values.

Let $\widehat Q_t^h(s,a)$ denote the sampled action value used at
horizon $h$ of iteration $t$. Define 
\begin{equation} \label{eq:backward-npg-continuation}
    f_t^h(s)
:=
\sum_{a\in\mathcal A}
\pi_{t+1}^h(a\mid s)\widehat Q_t^h(s,a).
\end{equation}
This is the sample-based value estimate computed using the updated policies $\pi_{t+1}^h,\ldots,\pi_{t+1}^H$. Both $f_t^h(s)$ and $\widehat Q_t^h(s,a)$ functions depend on
the policies and oracle observations; we suppress this dependence
for notational simplicity.
Set $\widehat Q_t^H(s,a)=R^H(s,a)$. For $h<H$, compute
\begin{equation} \label{eq:backward-npg-action-values}
    \widehat Q_t^h(s,a)
=
R^h(s,a)+(\bar P_t^h f_t^{h+1})(s,a),
\end{equation}
using the continuation estimate $f_t^{h+1}$ already computed
during the current iteration. At each horizon, update the
decision rule by
\begin{equation}
\label{eq:sample-based-npg-update}
\pi_{t+1}^h(a\mid s)
=
\frac{
    \pi_t^h(a\mid s)
    \exp\!\left(\eta_t\widehat Q_t^h(s,a)\right)
}{
    \displaystyle\sum_{a'\in\mathcal A}
    \pi_t^h(a'\mid s)
    \exp\!\left(\eta_t\widehat Q_t^h(s,a')\right)
},
\end{equation}
and then compute $f_t^h$ from its definition above.

Note that the update rule in \eqref{eq:backward-npg-action-values} is a sample-based variant of $Q^{\pi_{t+1},h}(s,a)
=
R^h(s,a)
+
(P^h V^{\pi_{t+1},h+1})(s,a)$ exact action value function.

For $h<H$, let $\mathcal G_t^{h+1}$ denote the information available
immediately before drawing the horizon-$h$ batches. It includes
$\mathcal F_t$ together with the observations and updates at horizons
$h+1,\ldots,H$ in the current sweep. Thus,
$\mathcal F_t\subseteq\mathcal G_t^{h+1}$. Since $f_t^{h+1}$ is
$\mathcal G_t^{h+1}$-measurable, fresh sampling gives
\[
\mathbb E\!\left[
    \widehat Q_t^h(s,a)
    \,\middle|\,
    \mathcal G_t^{h+1}
\right]
=
R^h(s,a)+(P^h f_t^{h+1})(s,a).
\]
This identity gives the conditional mean for the computed
continuation $f_t^{h+1}$; it does not establish unbiasedness
relative to the exact action value $Q^{\pi_{t+1},h}$.

Iteration $t$ uses $m_tSA(H-1)$ transition queries. For the returned policy $\pi$, performance is measured by the expected suboptimality gap under the initial distribution $\rho$.

\begin{rem}
\label{rem:sampled-continuation}
The model-based procedure of \citet{li2024breaking} estimates
transition probabilities from successor frequencies and performs
planning with the resulting estimated MDP. For comparison, the empirical transition probabilities
computed from the same batch would be
\[
\widehat P_t^h(s'\mid s,a)
:=
\frac{1}{m_t}
\sum_{\ell=1}^{m_t}
\mathbf 1\!\left\{Y_{t,\ell}^{h,s,a}=s'\right\},
\qquad s'\in\mathcal S.
\]
For every continuation function $v:\mathcal S\to\mathbb R$,
grouping the sampled values by successor state gives
\[
(\bar P_t^h v)(s,a)
=
\sum_{s'\in\mathcal S}
\widehat P_t^h(s'\mid s,a)v(s').
\]
This is an algebraic identity for the same batch of observations.

Our implementation computes the left-hand side with
$v=f_t^{h+1}$, which is available before the batch is drawn.
For each $(s,a)$, the agent accumulates
$f_t^{h+1}(Y_{t,\ell}^{h,s,a})$, discards each successor after
use, and divides the sum by $m_t$. It does not compute or
store $\widehat P_t^h(\cdot\mid s,a)$.
Thus, $\bar P_t^h$ denotes the sampled averaging operation,
not a stored transition matrix. The identity above shows that the two expressions give the
same scalar value; computing the sample average does not
require computing or storing the empirical transition
probabilities.
\end{rem}

\subsection{NPG with geometrically increasing step size}
\label{subsec:npg-doubling-batches}

\begin{algorithm}[t]
\FloatBarrier
\caption{NPG with doubling batches}
\label{alg:npg-doubling-batches}
\begin{algorithmic}[1]
\REQUIRE Generative access, known rewards, and number of
         iterations $T\ge1$.
\STATE Initialize $\pi_0^h(a\mid s)=1/A$ for every $(h,s,a)$.
\FOR{$t=0,\ldots,T-1$}
    \STATE Set $m_t=2^t$ and $\eta_t=m_t/H$.
    \FOR{$h=H,H-1,\ldots,1$}
        \IF{$h=H$}
            \STATE Set $\widehat Q_t^H=R^H$.
        \ELSE
            \STATE For each $(s,a)$, draw fresh successors
                   $Y_{t,\ell}^{h,s,a}\sim P^h(\cdot\mid s,a)$,
                   $\ell=1,\ldots,m_t$, and compute
                   $\widehat Q_t^h(s,a)$ by
                   \eqref{eq:backward-npg-action-values}.
        \ENDIF
        \STATE Update $\pi_{t+1}^h(\cdot\mid s)$ for every $s$
               by \eqref{eq:sample-based-npg-update}.
        \STATE Compute $f_t^h(s)$ for every $s$ by
               \eqref{eq:backward-npg-continuation}.
    \ENDFOR
\ENDFOR
\RETURN $\pi_T$.
\end{algorithmic}
\end{algorithm}
Algorithm~\ref{alg:npg-doubling-batches} uses batch size $m_t=2^t$ and step size $\eta_t=m_t/H$.
The schedules depend only on the iteration
index, so a single run can specify $\pi_T$ for every deterministic
$T\ge1$. After $T$ iterations, the total number of transition
queries is
\[
N
=
SA(H-1)\sum_{t=0}^{T-1}m_t
=
SA(H-1)(2^T-1).
\]

\begin{thm}[NPG with doubling batches]
\label{thm:npg-doubling-batches}
Consider an MDP satisfying the assumptions of
Section~\ref{subsec:finite-horizon-mdp}, and fix
$\rho\in\Delta(\mathcal S)$. For every deterministic integer
$T\ge1$, the policy $\pi_T$ returned by
Algorithm~\ref{alg:npg-doubling-batches} satisfies
\begin{equation}
\label{eq:npg-doubling-gap}
\mathbb E\!\left[
    V^{\star,1}(\rho)-V^{\pi_T,1}(\rho)
\right]
\le
\min\!\left\{
    H,\;
    21\sqrt{\frac{H^3\log(2SA)}{2^{T-1}}}
\right\}.
\end{equation}
\end{thm}

\begin{cor}[Sample complexity]
\label{cor:npg-doubling-sample-complexity}
Under the assumptions of Theorem~\ref{thm:npg-doubling-batches},
let $0<\varepsilon<H$ and choose
\[
T
=
1+
\left\lceil
\log_2\!\left(
    \frac{441H^3\log(2SA)}{\varepsilon^2}
\right)
\right\rceil.
\]
Then Algorithm~\ref{alg:npg-doubling-batches} returns a policy
$\pi_T$ satisfying $\mathbb E\!\left[
    V^{\star,1}(\rho)-V^{\pi_T,1}(\rho)
\right]
\le\varepsilon,$ using
\[
N
=
SA(H-1)(2^T-1)
<
\frac{1764SAH^4\log(2SA)}{\varepsilon^2}
\]
transition queries.
\end{cor}
\begin{proof}
The prescribed value of $T$ gives $2^{T-1}
\ge
\frac{441H^3\log(2SA)}{\varepsilon^2},$ so Theorem~\ref{thm:npg-doubling-batches} yields the expected
suboptimality-gap guarantee. Since
$2^{\lceil\log_2 x\rceil}<2x$ for every $x>0$, we have $2^T
<
\frac{1764H^3\log(2SA)}{\varepsilon^2}.$ Consequently, $N
=
SA(H-1)(2^T-1)
<
\frac{1764SA(H-1)H^3\log(2SA)}{\varepsilon^2}
<
\frac{1764SAH^4\log(2SA)}{\varepsilon^2}.$
\end{proof}

\subsection{NPG with constant step size}
\label{subsec:npg-constant-step}
We use Algorithm~\ref{alg:npg-doubling-batches} with
$m_t=m$ and $\eta_t=\eta$ for every $t=0,\ldots,T-1$,
where $m\ge2$ is an integer and $0<\eta\le1/(4H^2)$.
The initialization, updates, and output policy $\pi_T$ remain
the same. The step size is chosen before the first iteration
and can depend on $T$. The total number of transition queries is $N=SA(H-1)mT.$

\begin{thm}[NPG with constant step size]
\label{thm:npg-constant-step}
Consider an MDP as defined in
Section~\ref{subsec:finite-horizon-mdp}, and fix
$\rho\in\Delta(\mathcal S)$. For every deterministic integer
$T\ge1$, integer $m\ge2$, and constant step size
$0<\eta\le1/(4H^2)$, the policy $\pi_T$  satisfies
\begin{equation}
\label{eq:npg-constant-step-gap}
\mathbb E\!\left[
    V^{\star,1}(\rho)-V^{\pi_T,1}(\rho)
\right]
\le
\min\!\left\{
    H,\;
    \frac{H\log A}{\eta T}
    +
    \frac{30\eta H^2}{m}
    \sum_{t=1}^{T}\frac1t
\right\}.
\end{equation}
\end{thm}

\begin{cor}[Sample complexity]
\label{cor:npg-constant-step-sample-complexity}
Under the assumptions of Theorem~\ref{thm:npg-constant-step},
let $A\ge2$ and $0<\varepsilon\le1$. Choose $m=2$ and
an integer $T\ge1$ such that
\[
\frac{T}{1+\log T}
\ge
\frac{60H^3\log A}{\varepsilon^2},
\qquad
\eta
=
\sqrt{\frac{\log A}{15HT(1+\log T)}}.
\]
The constant step procedure returns an $\varepsilon$-optimal
policy in expectation under the initial distribution $\rho$,
using exactly $N=2SA(H-1)T$ transition queries. For the smallest such $T$,
\[
N
=
O\!\left(
    \frac{SAH^4\log A}{\varepsilon^2}
    \log\!\left(\frac{H^3\log A}{\varepsilon^2}\right)
\right).
\]
\end{cor}

\begin{proof}
The condition on $T$ implies
$T(1+\log T)\ge(16/15)H^3\log A$, ensuring
$0<\eta\le1/(4H^2)$. Theorem~\ref{thm:npg-constant-step}
and $\sum_{t=1}^{T}1/t\le1+\log T$ therefore give
\[
\begin{aligned}
\mathbb E\!\left[
    V^{\star,1}(\rho)-V^{\pi_T,1}(\rho)
\right]
&\le
\frac{H\log A}{\eta T}
+
15\eta H^2(1+\log T)
\\
&=
2\sqrt{\frac{15H^3\log A\,(1+\log T)}{T}}
\le\varepsilon.
\end{aligned}
\]
For the query bound, we write $C:=\frac{60H^3\log A}{\varepsilon^2},$\; $L:=1+\log(2C).$ Since $C>1$ and $L>1$,
\[
1+\log(2CL)=L+\log L\le2L.
\]
Thus $2CL/(1+\log(2CL))\ge C$. Monotonicity of
$x/(1+\log x)$ on $[1,\infty)$ shows that
$T=\lceil2CL\rceil$ satisfies the required condition.
The smallest admissible $T$ is consequently $O(C\log C)$,
which yields the stated bound upon substitution into
$N=2SA(H-1)T$.
\end{proof}

\subsection{NPG with a linearly decreasing step size}
\label{subsec:npg-linear-step}

We use Algorithm~\ref{alg:npg-doubling-batches} with $m_t=2$,\; $\eta_t=\eta_{\max}\frac{T-t}{T},$ where $T\ge1$ and $0<\eta_{\max}\le1/(4H^2)$ are chosen
before sampling. The initialization, updates, and output policy
$\pi_T$ remain the same. The step size decreases from
$\eta_{\max}$ to $\eta_{\max}/T$, and the total number of
transition queries is
\[
N=2SA(H-1)T.
\]

\begin{thm}[NPG with a linearly decreasing step size]
\label{thm:npg-linear-step}
Consider an MDP as defined in
Section~\ref{subsec:finite-horizon-mdp}, and fix
$\rho\in\Delta(\mathcal S)$. For every deterministic integer
$T\ge1$ and $0<\eta_{\max}\le1/(4H^2)$, the policy $\pi_T$
returned by the procedure above satisfies
\begin{equation}
\label{eq:npg-linear-step-gap}
\mathbb E\!\left[
    V^{\star,1}(\rho)-V^{\pi_T,1}(\rho)
\right]
\le
\min\!\left\{
    H,\;
    \frac{2H\log A}{\eta_{\max}(T+1)}
    +
    30\eta_{\max}H^2
\right\}.
\end{equation}
\end{thm}

\begin{cor}[Sample complexity]
\label{cor:npg-linear-step-sample-complexity}
Under the assumptions of Theorem~\ref{thm:npg-linear-step},
let $A\ge2$ and $0<\varepsilon\le1$. Choose
\[
T
=
\left\lceil
\frac{240H^3\log A}{\varepsilon^2}
\right\rceil,
\qquad
\eta_{\max}
=
\sqrt{\frac{\log A}{15H(T+1)}}.
\]
The procedure returns an $\varepsilon$-optimal policy in
expectation under the initial distribution $\rho$, using
\[
N
=
2SA(H-1)
\left\lceil
\frac{240H^3\log A}{\varepsilon^2}
\right\rceil
=
O\!\left(
\frac{SAH^4\log A}{\varepsilon^2}
\right)
\]
transition queries.
\end{cor}

\begin{proof}
The prescribed iteration count satisfies $T+1\ge\frac{16}{15}H^3\log A,$ so $0<\eta_{\max}\le1/(4H^2)$. Substituting the selected
$\eta_{\max}$ into Theorem~\ref{thm:npg-linear-step} gives
\[
\begin{aligned}
\mathbb E\!\left[
    V^{\star,1}(\rho)-V^{\pi_T,1}(\rho)
\right]
&\le
\frac{2H\log A}{\eta_{\max}(T+1)}
+
30\eta_{\max}H^2
\\
&=
4\sqrt{\frac{15H^3\log A}{T+1}}
\le\varepsilon.
\end{aligned}
\]
The query bound follows from using two samples per nonterminal
state-action-horizon triple at each iteration.
\end{proof}

\begin{rem}[The role of the step size]
The step size determines how strongly each NPG update favors actions
with larger estimated action values. A larger step size amplifies
differences between these estimates, including those due to sampling
error. The doubling schedule doubles both the step size and the batch
size at each iteration, reducing sampling error as the update becomes
more sensitive to it. The constant and linearly decreasing step size
schedules each use $m_t=2$ batches per nonterminal
state-action-horizon triple per sweep.
\end{rem}

\begin{rem}
When $\varepsilon$ is specified in advance,
a single backward sweep suffices. Under uniform policy initialization
and for $A\ge2$, choosing
\[
\eta_0=\frac{2H\log A}{\varepsilon},
\qquad
m_0=\left\lceil
\frac{144H^3\log(2SA)}{\varepsilon^2}
\right\rceil
\]
yields a policy $\pi_1$ that is $\varepsilon$-optimal in expectation
under the initial distribution $\rho$, using $SA(H-1)m_0$ transition
queries. This guarantee follows from
Lemma~\ref{lem:npg-weighted-comparison} with $T=1$ and
Lemma~\ref{lem:conditional-sampling-error} with $t=0$.
\end{rem}

\subsection{Information-theoretic lower bound}
\label{subsec:information-theoretic-lower-bound}

We establish a necessary transition-query budget for achieving
an expected suboptimality gap of at most $\varepsilon$ uniformly
over the family constructed in
Section~\ref{subsec:common-mdp-family}.
For this result, set $K=S-1$ and choose
\[
\rho(s_{i'})=\frac1K
\quad(i'\in[K]),
\qquad
\rho(g)=0.
\]
This choice ensures that errors at every decision state contribute
to the performance criterion, including when $H=2$, where
conditional uniform mixing cannot introduce another decision
before the terminal reward. An arbitrary initial distribution
could concentrate on one decision state and eliminate the
dependence on $S$ in that case. Uniformity is imposed only
for this lower-bound result.

\begin{thm}[Information-theoretic lower bound]
\label{thm:common-information-lower-bound}
There exists a universal constant $c>0$ such that, for all
integers $S,A,H\ge2$ and $N\ge1$, the following holds.
For every agent that uses at most $N$ transition queries
almost surely under the generative access model and returns
a Markov policy $\pi$, there exists an instance
$M_\theta\in\mathcal M_{K,A,H}$ such that
\begin{equation}
\label{eq:common-information-lower-bound}
\mathbb E_{M_\theta}\!\left[
    V^{\star,1}(\rho)-V^{\pi,1}(\rho)
\right]
\ge
c\min\!\left\{
    H,\;
    H^2\sqrt{\frac{SA}{N}}
\right\}.
\end{equation}
\end{thm}

The value functions are evaluated in $M_\theta$, and the
expectation includes the oracle observations and the agent's
internal randomness. The theorem permits adaptive query
selection and imposes no memory restriction. All instances
share the specified initial distribution and reward functions;
only their transition kernels vary. The proof appears in
Appendix~\ref{app:information-theoretic-lower-bound}.

\begin{cor}[Necessary transition-query budget]
\label{cor:necessary-transition-query-budget}
Let $0<\varepsilon<cH$, where $c$ is the constant in
Theorem~\ref{thm:common-information-lower-bound}.
Any agent satisfying
\[
\sup_{M_\theta\in\mathcal M_{K,A,H}}
\mathbb E_{M_\theta}\!\left[
    V^{\star,1}(\rho)-V^{\pi,1}(\rho)
\right]
\le\varepsilon
\]
must use a query budget satisfying
\[
N\ge\frac{c^2SAH^4}{\varepsilon^2}.
\]
\end{cor}

\begin{proof}
Since $\varepsilon<cH$,
Theorem~\ref{thm:common-information-lower-bound} implies $cH^2\sqrt{\frac{SA}{N}}\le\varepsilon.$ Rearranging proves the claim.
\end{proof}

Because $\mathcal M_{K,A,H}$ is a subclass of the finite-horizon
MDPs in Section~\ref{subsec:finite-horizon-mdp}, this necessary
budget also applies to any agent that guarantees the same
accuracy over the full class.

\begin{rem}(Significance of the matching bounds)
The information-theoretic lower bound shows that the proposed
NPG procedures achieve the optimal worst-case polynomial
dependence on $S$, $A$, $H$, and $1/\varepsilon$ under the
stated oracle and expected-gap criterion. This comparison
includes algorithms that estimate transition probabilities,
retain the full observation history, or select queries
adaptively. Thus, the optimal polynomial dependence is
attainable by the last NPG iterate using $O(SAH)$ memory
without estimating transition probabilities.
\end{rem}

\section{Conclusion}
\label{sec:conclusion}

To the best of our knowledge, this work provides the first
near-minimax-optimal sample-complexity guarantees for the last
iterate of NPG in finite-horizon tabular MDPs with
horizon-dependent transition kernels under generative access
and the expected optimality-gap criterion. We analyze three
batch and step size schedules, each attaining
$\widetilde O(SAH^4/\varepsilon^2)$ transition-query complexity
using $\mathcal{O}(SAH)$ memory without estimating transition
probabilities. The information-theoretic lower bound
establishes that this polynomial dependence is unavoidable,
even with adaptive query selection and unrestricted memory.
In the exact setting, our iteration lower bound also matches
the horizon and accuracy dependence of the corresponding
constant step size upper bound.

A natural next direction is to extend these guarantees to the infinite-horizon setting. Under generative access, an open question is whether the last-iterate sample-complexity bounds for NPG can be sharpened, both in their logarithmic dependence on the problem parameters and in their absolute constants.

\bibliographystyle{plainnat}
\bibliography{reference}

\appendix
\section*{Appendix}

\section{Properties of the constructed MDP family}
\label{app:family-properties}

Throughout this section, we fix an instance $M_\theta\in\mathcal M_{K,A,H}$
with specified admissible transition parameters and label array.

For a policy $\pi$, define
\[
\overline V^{\pi,h}
:=
\frac1K\sum_{i'=1}^{K}V^{\pi,h}(s_{i'}),
\qquad h\in[H],
\]
and
\[
\overline q^{\pi,h}
:=
\frac1K\sum_{i'=1}^{K}\sum_{k=1}^{A}
\pi^h(a_k\mid s_{i'})p_{\theta,i',a_k}^{h},
\qquad h\in[H-1].
\]
These are uniform averages over the decision states, irrespective
of $\rho$. The Bellman equations and the transition structure give
\begin{equation}
\label{eq:family-action-value}
Q^{\pi,h}(s_{i'},a_k)
=
p_{\theta,i',a_k}^{h}(H-h)
+
\bigl(1-p_{\theta,i',a_k}^{h}\bigr)
\overline V^{\pi,h+1},
\end{equation}
for $h\in[H-1]$, $i'\in[K]$, and $k\in[A]$.
Consequently,
\begin{equation}
\label{eq:averaged-value-recursion}
\overline V^{\pi,h}
=
\overline q^{\pi,h}(H-h)
+
\bigl(1-\overline q^{\pi,h}\bigr)
\overline V^{\pi,h+1},
\end{equation}
with $\overline V^{\pi,H}=0$.
Whenever $\theta_{i'}^h\ne0$, subtracting the action values gives
\begin{equation}
\label{eq:family-action-gap-identity}
Q^{\pi,h}(s_{i'},a_{\theta_{i'}^h})
-
Q^{\pi,h}(s_{i'},a_k)
=
2\alpha^h
\left[(H-h)-\overline V^{\pi,h+1}\right],
\qquad
k\in[A]\setminus\{\theta_{i'}^h\}.
\end{equation}

\begin{lem}
\label{lem:family-action-values}
For every policy $\pi$ and $h\in[H-1]$,
\begin{equation}
\label{eq:family-continuation-bound}
0
\le
\overline V^{\pi,h+1}
\le
\frac{(H-h-1)(H-h)}{4H}
\le
\frac{H-h}{4}.
\end{equation}
For every $i'\in[K]$ with $\theta_{i'}^h\ne0$ and every
$k\in[A]\setminus\{\theta_{i'}^h\}$,
\begin{equation}
\label{eq:family-action-gap}
\frac32\alpha^h(H-h)
\le
Q^{\pi,h}(s_{i'},a_{\theta_{i'}^h})
-
Q^{\pi,h}(s_{i'},a_k)
\le
2\alpha^h(H-h).
\end{equation}
Consequently, the favorable action uniquely maximizes the
action value under every continuation policy, and every optimal
policy satisfies
\[
\pi^{\star,h}(a_{\theta_{i'}^h}\mid s_{i'})=1
\]
whenever $h\in[H-1]$ and $\theta_{i'}^h\ne0$.
\end{lem}

\begin{proof}
By \eqref{eq:transition-parameter-restrictions}, the probability
of entering $g$ from any decision state under any action is at
most $1/(2H)$.

Starting from $s_{i'}$ at horizon $h+1$, entry into $g$ before
horizon $i\in\{h+1,\ldots,H\}$ must occur during one of the
preceding $i-h-1$ transitions. Since $g$ is absorbing, these
entry events are disjoint, and
\[
\begin{aligned}
d_{s_{i'}}^{h+1\to i,\pi}(g)
&=
\sum_{j=h+1}^{i-1}
\mathbb P\!\left(
S^j\ne g,\ S^{j+1}=g
\,\middle|\,
S^{h+1}=s_{i'}
\right)\\
&\le
\frac{i-h-1}{2H}.
\end{aligned}
\]
The sum is empty when $i=h+1$. Therefore,
\[
\begin{aligned}
V^{\pi,h+1}(s_{i'})
&=
\sum_{i=h+1}^{H}
d_{s_{i'}}^{h+1\to i,\pi}(g)\\
&\le
\frac1{2H}\sum_{i=h+1}^{H}(i-h-1)\\
&=
\frac{(H-h-1)(H-h)}{4H}.
\end{aligned}
\]
Averaging over decision states, using nonnegative rewards and
$H-h-1\le H$, proves \eqref{eq:family-continuation-bound}.

For $\theta_{i'}^h\ne0$, this bound implies
\[
\frac34(H-h)
\le
(H-h)-\overline V^{\pi,h+1}
\le
H-h.
\]
Substituting into \eqref{eq:family-action-gap-identity} proves
\eqref{eq:family-action-gap}. Its lower bound is strictly positive
because $\alpha^h>0$ and $h<H$.

Applying this conclusion under an optimal continuation policy
shows that the favorable action uniquely maximizes
$Q^{\star,h}(s_{i'},\cdot)$. Bellman optimality then requires
$\pi^{\star,h}(a_{\theta_{i'}^h}\mid s_{i'})=1$.
\end{proof}

For $h\in[H-1]$ and $i'\in[K]$, define
\[
\mathrm{w}_\pi^h(i')
:=
\begin{cases}
0,
& \theta_{i'}^h=0,\\[1mm]
1-\pi^h(a_{\theta_{i'}^h}\mid s_{i'}),
& \theta_{i'}^h\ne0,
\end{cases}
\qquad
\overline w_\pi^h
:=
\frac1K\sum_{i'=1}^{K}\mathrm{w}_\pi^h(i').
\]
Thus, $\mathrm{w}_\pi^h(i')$ is the probability of selecting a suboptimal
action at $(h,s_{i'})$.
For $h\in[H]$ and $s\in\mathcal S$, write
\[
d_\rho^{1\to h,\star}(s)
:=
\sum_{\widetilde s\in\mathcal S}
\rho(\widetilde s)
d_{\widetilde s}^{1\to h,\pi^\star}(s).
\]

\begin{lem}
\label{lem:family-policy-loss}
For every policy $\pi$,
\begin{equation}
\label{eq:policy-loss-identity}
\begin{aligned}
V^{\star,1}(\rho)-V^{\pi,1}(\rho)
={}&
\sum_{h=1}^{H-1}
2\alpha^h
\left[(H-h)-\overline V^{\pi,h+1}\right]
\sum_{i'=1}^{K}
d_\rho^{1\to h,\star}(s_{i'})
\mathrm{w}_\pi^h(i').
\end{aligned}
\end{equation}
Moreover,
\begin{equation}
\label{eq:policy-loss-lower-bound}
\begin{aligned}
V^{\star,1}(\rho)-V^{\pi,1}(\rho)
\ge{}
\frac32\alpha^1(H-1)
\sum_{i'=1}^{K}\rho(s_{i'})w_\pi^1(i') +
\frac34\sum_{h=2}^{H-1}
\alpha^h(H-h)\overline w_\pi^h.
\end{aligned}
\end{equation}
The second sum is empty when $H=2$.
\end{lem}

\begin{proof}
Fix a policy $\pi$. At a decision state with a nonzero label,
Lemma~\ref{lem:family-action-values} implies that an optimal policy
selects the favorable action. Hence,
\[
\begin{aligned}
\sum_{a\in\mathcal A}
\pi^{\star,h}(a\mid s_{i'})
Q^{\pi,h}(s_{i'},a)
-
V^{\pi,h}(s_{i'}) =
2\alpha^h
\left[(H-h)-\overline V^{\pi,h+1}\right]
\mathrm{w}_\pi^h(i').
\end{aligned}
\]
The same equality holds when the label is zero, because both
sides vanish. The corresponding difference is also zero at $g$
and at horizon $H$.

For $h\in[H-1]$, subtracting the Bellman equations gives
\[
\begin{aligned}
V^{\star,h}(s)-V^{\pi,h}(s)
={}&
\sum_{a\in\mathcal A}
\pi^{\star,h}(a\mid s)
Q^{\pi,h}(s,a)
-
V^{\pi,h}(s)\\
&+
\sum_{a\in\mathcal A}\sum_{s'\in\mathcal S}
\pi^{\star,h}(a\mid s)
P_\theta^h(s'\mid s,a)
\left[
V^{\star,h+1}(s')-V^{\pi,h+1}(s')
\right].
\end{aligned}
\]
Since $V^{\star,H}(s)=V^{\pi,H}(s)$ for every state in this
family, iterating this identity and averaging the initial state
under $\rho$ yields
\[
\begin{aligned}
V^{\star,1}(\rho)-V^{\pi,1}(\rho)
=
\sum_{h=1}^{H-1}\sum_{s\in\mathcal S}
d_\rho^{1\to h,\star}(s)
\left[
\sum_{a\in\mathcal A}
\pi^{\star,h}(a\mid s)Q^{\pi,h}(s,a)
-
V^{\pi,h}(s)
\right].
\end{aligned}
\]
Substituting the expression for the local difference proves
\eqref{eq:policy-loss-identity}.

At horizon $1$,
\[
d_\rho^{1\to1,\star}(s_{i'})=\rho(s_{i'}).
\]
For $h\in\{2,\ldots,H\}$, conditional uniform mixing gives
\[
d_\rho^{1\to h,\star}(s_{i'})
=
\frac1K\sum_{j'=1}^{K}
d_\rho^{1\to h,\star}(s_{j'}).
\]
Since $\rho(g)=0$ and the probability of entering $g$ at each
transition is at most $1/(2H)$,
\[
\begin{aligned}
\sum_{j'=1}^{K}
d_\rho^{1\to h,\star}(s_{j'})
&\ge
\left(1-\frac1{2H}\right)^{h-1}\\
&\ge
1-\frac{h-1}{2H}\\
&\ge
\frac12,
\end{aligned}
\]
where the second inequality is Bernoulli's inequality.
Consequently,
\[
d_\rho^{1\to h,\star}(s_{i'})
\ge
\frac1{2K},
\qquad h\in\{2,\ldots,H\}.
\]

By Lemma~\ref{lem:family-action-values},
\[
2\alpha^h
\left[(H-h)-\overline V^{\pi,h+1}\right]
\ge
\frac32\alpha^h(H-h).
\]
Combining this inequality with the initial visitation probabilities
at $h=1$ and the visitation lower bound for $h\ge2$ proves
\eqref{eq:policy-loss-lower-bound}.
\end{proof}

\subsection{Proof of the iteration lower bound for exact NPG}
\label{app:exact-npg-lower-bound}

\begin{proof}[Proof of Theorem~\ref{thm:exact-npg-lower-bound}]
For each $h\in[H-1]$, set
\[
p^h=\frac{1}{4H},
\qquad
\alpha^h=\alpha:=
\frac{1}{8H}
\min\!\left\{1,\frac{\log A}{\eta T}\right\}.
\]
Choose any label array with $\theta_{i'}^h\in[A]$ for every
$h\in[H-1]$ and $i'\in[K]$, so that each decision state has a
favorable action at every nonterminal horizon. Since
$0<\alpha<p^h$ and $p^h+\alpha\le 3/(8H)\le 1/(2H)$,
the resulting instance belongs to $\mathcal M_{K,A,H}$.

Fix $h\in[H-1]$ and $i'\in[K]$, and write
$a^\star:=a_{\theta_{i'}^h}$ for the favorable action at
$(h,s_{i'})$. For each $k\in[A]\setminus\{\theta_{i'}^h\}$,
uniform initialization and the exact NPG update
\eqref{eq:npg-update} give
\[
\begin{aligned}
\log\frac{\pi_T^h(a^\star\mid s_{i'})}
          {\pi_T^h(a_k\mid s_{i'})}
&=
\eta\sum_{t=0}^{T-1}
\left[
Q^{\pi_t,h}(s_{i'},a^\star)
-
Q^{\pi_t,h}(s_{i'},a_k)
\right] \\
&\le 2\eta T\alpha(H-h)
\le \frac14\min\{\eta T,\log A\}
\le \frac{\log A}{4},
\end{aligned}
\]
where the first inequality follows from \eqref{eq:family-action-gap}.
Thus, each unfavorable action has probability at least
$A^{-1/4}\pi_T^h(a^\star\mid s_{i'})$. Summing over the $A-1$
unfavorable actions yields
\[
\mathrm{w}_{\pi_T}^h(i')
\ge
(A-1)A^{-1/4}\bigl(1-\mathrm{w}_{\pi_T}^h(i')\bigr).
\]
It follows that
\[
\mathrm{w}_{\pi_T}^h(i')
\ge
\frac{A-1}{A-1+A^{1/4}}
>
\frac13,
\]
where the strict inequality uses $A^{1/4}<A\le 2(A-1)$ for $A\ge 2$.

This bound holds at every decision state and every nonterminal
horizon. Hence, for any $\rho\in\Delta(\mathcal S)$ with $\rho(g)=0$,
\[
\sum_{i'=1}^{K}\rho(s_{i'})\mathrm{w}_{\pi_T}^1(i')
\ge \frac13,
\qquad
\overline w_{\pi_T}^h\ge\frac13
\quad\text{for every }h\in[H-1].
\]
Applying \eqref{eq:policy-loss-lower-bound} gives
\[
\begin{aligned}
V^{\star,1}(\rho)-V^{\pi_T,1}(\rho)
&\ge
\frac{\alpha}{2}(H-1)
+
\frac{\alpha}{4}\sum_{h=2}^{H-1}(H-h) \\
&\ge
\frac{\alpha}{4}\sum_{h=1}^{H-1}(H-h)
=
\frac{\alpha H(H-1)}{8} \\
&=
\frac{H-1}{64}
\min\!\left\{1,\frac{\log A}{\eta T}\right\} \\
&\ge
\frac{1}{128}
\min\!\left\{H,\frac{H\log A}{\eta T}\right\},
\end{aligned}
\]
where the last inequality uses $H-1\ge H/2$.

For $\eta=1/H$, the lower bound becomes
$\frac{1}{128}\min\{H,H^2\log A/T\}$. Since
$\varepsilon<H/128$, a gap of at most $\varepsilon$ requires
$H^2\log A/(128T)\le\varepsilon$, which gives the stated
iteration lower bound.
\end{proof}

\section{Supporting lemmas for sample-based setting}
\label{app:sample-based-npg-lemmas}

Throughout this section, we consider the MDP formulation defined in Section~\ref{subsec:finite-horizon-mdp} and fix an initial distribution $\rho\in\Delta(\mathcal S)$. Consider the NPG update defined in Section~\ref{sec:sample-based}, with deterministic
positive integer batch sizes $m_t$ and finite deterministic step sizes $\eta_t>0$.

\begin{lem}
\label{lem:sampled-value-bounds}
For every iteration $t\ge0$, horizon $h\in[H]$,
state $s\in\mathcal S$, and action $a\in\mathcal A$,
\[
0\le\widehat Q_t^h(s,a)\le H-h+1,
\qquad
0\le f_t^h(s)\le H-h+1.
\]
\end{lem}

\begin{proof}
Fix $t$ and proceed by backward induction. At horizon $H$,
$\widehat Q_t^H=R^H\in[0,1]$, and $f_t^H(s)$ is a convex
combination of these action values.

Suppose $0\le f_t^{h+1}(s')\le H-h$ for every
$s'\in\mathcal S$. Then
\[
0\le(\bar P_t^h f_t^{h+1})(s,a)\le H-h.
\]
Since $R^h(s,a)\in[0,1]$,
\[
0\le
\widehat Q_t^h(s,a)
=
R^h(s,a)+(\bar P_t^h f_t^{h+1})(s,a)
\le H-h+1.
\]
The same bounds hold for $f_t^h(s)$, which is a convex
combination of $\widehat Q_t^h(s,\cdot)$.
\end{proof}

For a policy $\pi$, write
\[
d_\rho^{1\to h,\pi}(s)
:=
\sum_{\widetilde s\in\mathcal S}
\rho(\widetilde s)d_{\widetilde s}^{1\to h,\pi}(s).
\]
Let $w_t^h(s,a)$ denote the probability of visiting state $s$
and selecting action $a$ at horizon $h$, starting from $\rho$
and following the realized policy $\pi_{t+1}$ under the true
transition kernels. Thus,
\[
w_t^h(s,a)
:=
d_\rho^{1\to h,\pi_{t+1}}(s)\,
\pi_{t+1}^h(a\mid s).
\]
For each realization of $\pi_{t+1}$,
\[
w_t^h(s,a)\ge0,
\qquad
\sum_{s\in\mathcal S}\sum_{a\in\mathcal A}w_t^h(s,a)=1.
\]
These probabilities depend on the oracle observations through
$\pi_{t+1}$; they describe policy visitation rather than oracle
query frequencies.

For $h<H$, define the sampling error
\[
e_t^h(s,a)
:=
\bigl((\bar P_t^h-P^h)f_t^{h+1}\bigr)(s,a),
\]
and set $e_t^H=0$. Define its visitation-weighted absolute sum by
\[
\mathcal E_t
:=
\sum_{h=1}^{H}\sum_{s,a}
w_t^h(s,a)|e_t^h(s,a)|.
\]
Here and below, $\sum_{s,a}$ denotes summation over
$\mathcal S\times\mathcal A$. Also write
\[
f_t^1(\rho):=\sum_{s\in\mathcal S}\rho(s)f_t^1(s).
\]

\begin{lem}
\label{lem:sampled-value-error-identity}
For every iteration $t\ge0$ and every realization of the oracle
observations,
\[
f_t^1(\rho)-V^{\pi_{t+1},1}(\rho)
=
\sum_{h=1}^{H}\sum_{s,a}
w_t^h(s,a)e_t^h(s,a).
\]
Consequently,
\[
\left|f_t^1(\rho)-V^{\pi_{t+1},1}(\rho)\right|
\le\min\{H,\mathcal E_t\}.
\]
\end{lem}

\begin{proof}
Fix $t$ and a realization of the observations. For $h<H$,
the sampled-value recursion and the Bellman equation give
\[
\begin{aligned}
f_t^h(s)-V^{\pi_{t+1},h}(s)
=
\sum_a\pi_{t+1}^h(a\mid s)
\Bigl[
e_t^h(s,a)
+
\bigl(P^h(f_t^{h+1}-V^{\pi_{t+1},h+1})\bigr)(s,a)
\Bigr].
\end{aligned}
\]
Multiply by $d_\rho^{1\to h,\pi_{t+1}}(s)$ and sum over states.
The visitation recursion yields
\[
\begin{aligned}
\sum_s d_\rho^{1\to h,\pi_{t+1}}(s)
\bigl(f_t^h(s)-V^{\pi_{t+1},h}(s)\bigr)
&=
\sum_{s,a}w_t^h(s,a)e_t^h(s,a)
\\
&\qquad+
\sum_{s'}d_\rho^{1\to h+1,\pi_{t+1}}(s')
\bigl(f_t^{h+1}(s')-V^{\pi_{t+1},h+1}(s')\bigr).
\end{aligned}
\]
At horizon $H$, $f_t^H=V^{\pi_{t+1},H}$. Summing over
$h=1,\ldots,H-1$ therefore proves the identity.

The bound by $\mathcal E_t$ follows from the triangle inequality.
The bound by $H$ follows because both $f_t^1(\rho)$ and
$V^{\pi_{t+1},1}(\rho)$ belong to $[0,H]$.
\end{proof}

For $h<H$, define
\[
v_t^h(s,a)
:=
\bigl(P^h(f_t^{h+1})^2\bigr)(s,a)
-
\bigl((P^h f_t^{h+1})(s,a)\bigr)^2,
\]
and set $v_t^H=0$. Conditional on $\mathcal G_t^{h+1}$, this
is the variance of $f_t^{h+1}(Y)$ for a fresh successor
$Y\sim P^h(\cdot\mid s,a)$. Define the visitation-weighted
sum of these variances by
\[
W_t
:=
\sum_{h=1}^{H}\sum_{s,a}
w_t^h(s,a)v_t^h(s,a).
\]

\begin{lem}
\label{lem:weighted-sampling-error}
Fix an iteration $t\ge0$. For every constant
$\lambda\in(0,m_t/H]$,
\[
\mathbb E[\mathcal E_t\mid\mathcal F_t]
\le
\frac{\lambda}{m_t}\mathbb E[W_t\mid\mathcal F_t]
+
\frac{H\log(2SA)}{\lambda}
\qquad\text{almost surely}.
\]
\end{lem}

\begin{proof}
Fix $h<H$ and $(s,a)$. Conditional on $\mathcal G_t^{h+1}$,
the variables
\[
X_\ell
:=
f_t^{h+1}\!\left(Y_{t,\ell}^{h,s,a}\right)
-
(P^h f_t^{h+1})(s,a),
\qquad \ell=1,\ldots,m_t,
\]
are independent and satisfy
\[
\mathbb E[X_\ell\mid\mathcal G_t^{h+1}]=0,
\qquad
\mathbb E[X_\ell^2\mid\mathcal G_t^{h+1}]
=
v_t^h(s,a),
\qquad
|X_\ell|\le H.
\]
Moreover,
\[
e_t^h(s,a)=\frac1{m_t}\sum_{\ell=1}^{m_t}X_\ell.
\]

For $\tau\in\{-1,1\}$, the condition $\lambda H\le m_t$
and the inequality $e^x\le1+x+x^2$ for $|x|\le1$ imply
\[
\begin{aligned}
\mathbb E\!\left[
e^{\tau\lambda X_\ell/m_t}
\,\middle|\,
\mathcal G_t^{h+1}
\right]
&\le
1+\frac{\lambda^2}{m_t^2}v_t^h(s,a)
\\
&\le
\exp\!\left(
\frac{\lambda^2}{m_t^2}v_t^h(s,a)
\right).
\end{aligned}
\]
Conditional independence then gives
\[
\mathbb E\!\left[
\exp\!\left(
\tau\lambda e_t^h(s,a)
-
\frac{\lambda^2}{m_t}v_t^h(s,a)
\right)
\,\middle|\,
\mathcal G_t^{h+1}
\right]
\le1.
\]

Define
\[
\Psi_t^h
:=
\log\!\left[
\sum_{s,a}\sum_{\tau\in\{-1,1\}}
\exp\!\left(
\tau\lambda e_t^h(s,a)
-
\frac{\lambda^2}{m_t}v_t^h(s,a)
\right)
\right].
\]
For every realization and every $(s,a)$,
\[
\lambda|e_t^h(s,a)|
-
\frac{\lambda^2}{m_t}v_t^h(s,a)
\le\Psi_t^h.
\]
Multiplying by $w_t^h(s,a)$ and summing over state-action
pairs gives
\[
\lambda\sum_{s,a}w_t^h(s,a)|e_t^h(s,a)|
-
\frac{\lambda^2}{m_t}
\sum_{s,a}w_t^h(s,a)v_t^h(s,a)
\le\Psi_t^h.
\]
This inequality holds for every realization, including the
resulting sample-dependent visitation probabilities.

The exponential bound, the tower property, and conditional
Jensen inequality imply
\[
\mathbb E[\Psi_t^h\mid\mathcal F_t]
\le
\log\mathbb E[e^{\Psi_t^h}\mid\mathcal F_t]
\le\log(2SA).
\]
Taking conditional expectations in the weighted inequality,
summing over $h=1,\ldots,H-1$, and dividing by $\lambda$,
we obtain
\[
\mathbb E[\mathcal E_t\mid\mathcal F_t]
\le
\frac{\lambda}{m_t}\mathbb E[W_t\mid\mathcal F_t]
+
\frac{(H-1)\log(2SA)}{\lambda}.
\]
The stated bound follows from $H-1\le H$.
\end{proof}

\begin{lem}
\label{lem:sum-transition-variances}
For every iteration $t\ge0$ and every realization of the oracle
observations,
\[
W_t\le 2H(H+\mathcal E_t).
\]
\end{lem}

\begin{proof}
Fix $t$ and a realization of the observations. For $h<H$, define
\[
c_t^h(s)
:=
\sum_{a\in\mathcal A}
\pi_{t+1}^h(a\mid s)(P^h f_t^{h+1})(s,a),
\]
and set $c_t^H=0$. By Jensen's inequality,
\[
\sum_a\pi_{t+1}^h(a\mid s)
\bigl((P^h f_t^{h+1})(s,a)\bigr)^2
\ge \bigl(c_t^h(s)\bigr)^2.
\]
Combining this inequality with the visitation recursion gives
\[
\begin{aligned}
\sum_{s,a}w_t^h(s,a)v_t^h(s,a)
\le
\sum_{s'}d_\rho^{1\to h+1,\pi_{t+1}}(s')
\bigl(f_t^{h+1}(s')\bigr)^2 -
\sum_s d_\rho^{1\to h,\pi_{t+1}}(s)
\bigl(c_t^h(s)\bigr)^2.
\end{aligned}
\]
Summing over $h=1,\ldots,H-1$ yields
\[
\begin{aligned}
W_t
&\le
\sum_{h=1}^{H}\sum_s
d_\rho^{1\to h,\pi_{t+1}}(s)
\left[
\bigl(f_t^h(s)\bigr)^2-\bigl(c_t^h(s)\bigr)^2
\right]
-
\sum_s\rho(s)\bigl(f_t^1(s)\bigr)^2
\\
&\le
\sum_{h=1}^{H}\sum_s
d_\rho^{1\to h,\pi_{t+1}}(s)
\bigl(f_t^h(s)-c_t^h(s)\bigr)
\bigl(f_t^h(s)+c_t^h(s)\bigr).
\end{aligned}
\]
Lemma~\ref{lem:sampled-value-bounds} implies
\[
0\le f_t^h(s)+c_t^h(s)\le 2H.
\]
Moreover, the sampled-value recursion gives
\[
\begin{aligned}
f_t^h(s)-c_t^h(s)
&=
\sum_a\pi_{t+1}^h(a\mid s)
\bigl(R^h(s,a)+e_t^h(s,a)\bigr)
\\
&\le
1+\sum_a\pi_{t+1}^h(a\mid s)|e_t^h(s,a)|.
\end{aligned}
\]
If $f_t^h(s)-c_t^h(s)<0$, its product with
$f_t^h(s)+c_t^h(s)$ is nonpositive. Otherwise, that product
is at most
\[
2H\left(
1+\sum_a\pi_{t+1}^h(a\mid s)|e_t^h(s,a)|
\right).
\]
Consequently,
\[
\begin{aligned}
W_t
&\le
2H\sum_{h=1}^{H}\sum_s
d_\rho^{1\to h,\pi_{t+1}}(s)
\left(
1+\sum_a\pi_{t+1}^h(a\mid s)|e_t^h(s,a)|
\right)
\\
&=
2H(H+\mathcal E_t).
\end{aligned}
\]
\end{proof}

\begin{lem}
\label{lem:conditional-sampling-error}
For every iteration $t\ge0$,
\[
\mathbb E\!\left[
f_t^1(\rho)-V^{\pi_{t+1},1}(\rho)
\,\middle|\,
\mathcal F_t
\right]
\le
\min\!\left\{
H,\;
6\sqrt{\frac{H^3\log(2SA)}{m_t}}
\right\}
\qquad\text{almost surely}.
\]
\end{lem}

\begin{proof}
Lemma~\ref{lem:sampled-value-error-identity} gives the bound $H$.
If $m_t<16H\log(2SA)$, then
\[
6\sqrt{\frac{H^3\log(2SA)}{m_t}}>\frac{3H}{2},
\]
so the claimed bound follows.

Suppose $m_t\ge16H\log(2SA)$, and choose
\[
\lambda
=
\sqrt{\frac{m_t\log(2SA)}{H}}.
\]
Then
\[
\frac{\lambda H}{m_t}
=
\sqrt{\frac{H\log(2SA)}{m_t}}
\le\frac14,
\]
so Lemma~\ref{lem:weighted-sampling-error} applies. Combining
Lemmas~\ref{lem:weighted-sampling-error}
and~\ref{lem:sum-transition-variances} gives
\[
\mathbb E[\mathcal E_t\mid\mathcal F_t]
\le
\frac{2\lambda H^2}{m_t}
+
\frac{2\lambda H}{m_t}
\mathbb E[\mathcal E_t\mid\mathcal F_t]
+
\frac{H\log(2SA)}{\lambda}.
\]
The range bounds imply $\mathcal E_t\le H(H-1)$, so all terms
are finite. Rearranging,
\[
\begin{aligned}
\left(1-\frac{2\lambda H}{m_t}\right)
\mathbb E[\mathcal E_t\mid\mathcal F_t]
&\le
\frac{2\lambda H^2}{m_t}
+
\frac{H\log(2SA)}{\lambda}
\\
&=
3\sqrt{\frac{H^3\log(2SA)}{m_t}}.
\end{aligned}
\]
The coefficient on the left is at least $1/2$. Hence
\[
\mathbb E[\mathcal E_t\mid\mathcal F_t]
\le
6\sqrt{\frac{H^3\log(2SA)}{m_t}}.
\]
Apply Lemma~\ref{lem:sampled-value-error-identity} and combine
this bound with the bound $H$.
\end{proof}

For the next lemma, fix an optimal policy $\pi^\star$
independently of the oracle observations. Using the same samples
as iteration $t$, define
\[
\widetilde V_t^{\star,H}(s)
:=
\sum_a\pi^{\star,H}(a\mid s)R^H(s,a),
\]
and, for $h<H$,
\[
\widetilde V_t^{\star,h}(s)
:=
\sum_a\pi^{\star,h}(a\mid s)
\left[
R^h(s,a)
+
(\bar P_t^h\widetilde V_t^{\star,h+1})(s,a)
\right].
\]
These auxiliary quantities evaluate the fixed optimal policy
using the sampled transitions and are used only in the analysis.

\begin{lem}
\label{lem:optimal-policy-comparison}
For every iteration $t\ge0$, horizon $h\in[H]$, and state
$s\in\mathcal S$,
\[
\mathbb E\!\left[
\widetilde V_t^{\star,h}(s)
\,\middle|\,
\mathcal F_t
\right]
=
V^{\star,h}(s)
\qquad\text{almost surely}.
\]
Moreover, for every realization and every $h<H$,
\[
\begin{aligned}
\widetilde V_t^{\star,h}(s)-f_t^h(s) =
\sum_a
\bigl(\pi^{\star,h}(a\mid s)-\pi_{t+1}^h(a\mid s)\bigr)
\widehat Q_t^h(s,a) +
\sum_a\pi^{\star,h}(a\mid s)
\bigl(
\bar P_t^h(\widetilde V_t^{\star,h+1}-f_t^{h+1})
\bigr)(s,a).
\end{aligned}
\]
At horizon $H$, the identity holds with the continuation term
omitted.
\end{lem}

\begin{proof}
At horizon $H$,
\[
\widetilde V_t^{\star,H}(s)=V^{\star,H}(s).
\]
For $h<H$, the function $\widetilde V_t^{\star,h+1}$ depends
only on samples at later horizons and is therefore
$\mathcal G_t^{h+1}$-measurable. Fresh sampling gives
\[
\mathbb E\!\left[
(\bar P_t^h\widetilde V_t^{\star,h+1})(s,a)
\,\middle|\,
\mathcal G_t^{h+1}
\right]
=
(P^h\widetilde V_t^{\star,h+1})(s,a).
\]
Taking conditional expectations given $\mathcal F_t$,
\[
\begin{aligned}
\mathbb E\!\left[
(\bar P_t^h\widetilde V_t^{\star,h+1})(s,a)
\,\middle|\,
\mathcal F_t
\right] =
\sum_{s'}P^h(s'\mid s,a)
\mathbb E\!\left[
\widetilde V_t^{\star,h+1}(s')
\,\middle|\,
\mathcal F_t
\right].
\end{aligned}
\]
Backward induction and the Bellman equation for $\pi^\star$
establish the first assertion.

For the second assertion, write
\[
\begin{aligned}
\widetilde V_t^{\star,h}(s)
=
\sum_a\pi^{\star,h}(a\mid s)
\left[
R^h(s,a)+(\bar P_t^h f_t^{h+1})(s,a)
\right] +
\sum_a\pi^{\star,h}(a\mid s)
\bigl(
\bar P_t^h(\widetilde V_t^{\star,h+1}-f_t^{h+1})
\bigr)(s,a).
\end{aligned}
\]
The first sum equals
$\sum_a\pi^{\star,h}(a\mid s)\widehat Q_t^h(s,a)$.
Subtracting
\[
f_t^h(s)
=
\sum_a\pi_{t+1}^h(a\mid s)\widehat Q_t^h(s,a)
\]
proves the identity. At $H$, it follows directly from
$\widehat Q_t^H=R^H$.
\end{proof}

For distributions $p,q\in\Delta(\mathcal A)$ with $q(a)>0$
for every action, define
\[
D_{\mathrm{KL}}(p\|q)
:=
\sum_{a\in\mathcal A}p(a)\log\frac{p(a)}{q(a)},
\]
with zero-probability terms interpreted as zero.

\begin{lem}[Three-point identity]
\label{lem:npg-three-point-identity}
Let $u\in\Delta(\mathcal A)$ satisfy $u(a)>0$ for every
$a\in\mathcal A$, let $z:\mathcal A\to\mathbb R$, and let
$\eta>0$. Define
\[
u^+(a)
:=
\frac{u(a)e^{\eta z(a)}}
{\sum_{a'\in\mathcal A}u(a')e^{\eta z(a')}}.
\]
Then, for every $\nu\in\Delta(\mathcal A)$,
\[
\begin{aligned}
\eta\sum_a\bigl(\nu(a)-u^+(a)\bigr)z(a)
&=
D_{\mathrm{KL}}(\nu\|u)
-
D_{\mathrm{KL}}(\nu\|u^+)
-
D_{\mathrm{KL}}(u^+\|u)
\\
&\le
D_{\mathrm{KL}}(\nu\|u)
-
D_{\mathrm{KL}}(\nu\|u^+).
\end{aligned}
\]
\end{lem}

\begin{proof}
The definition of $u^+$ gives
\[
\log\frac{u^+(a)}{u(a)}
=
\eta z(a)
-
\log\!\left(
\sum_{a'}u(a')e^{\eta z(a')}
\right).
\]
Since $\sum_a(\nu(a)-u^+(a))=0$,
\[
\eta\sum_a(\nu(a)-u^+(a))z(a)
=
\sum_a(\nu(a)-u^+(a))
\log\frac{u^+(a)}{u(a)}.
\]
Expanding the right side proves the identity. The inequality
follows from $D_{\mathrm{KL}}(u^+\|u)\ge0$.
\end{proof}

\begin{lem}
\label{lem:npg-weighted-comparison}
Under uniform initialization, for every integer $T\ge1$,
\[
\sum_{t=0}^{T-1}\eta_t
\mathbb E\!\left[
V^{\star,1}(\rho)-f_t^1(\rho)
\right]
\le H\log A.
\]
\end{lem}

\begin{proof}
Define
\[
\kappa_t^h(s)
:=
D_{\mathrm{KL}}\!\left(
\pi^{\star,h}(\cdot\mid s)
\middle\|
\pi_t^h(\cdot\mid s)
\right),
\]
and
\[
\Phi_t
:=
\sum_{h=1}^{H}\sum_s
d_\rho^{1\to h,\pi^\star}(s)\kappa_t^h(s).
\]
The update preserves full support. Moreover,
Lemma~\ref{lem:sampled-value-bounds} and the multiplicative
update imply
\[
-\log\pi_t^h(a\mid s)
\le
\log A+H\sum_{t'=0}^{t-1}\eta_{t'}.
\]
Thus, the KL terms are finite and integrable for every finite
$t$. Uniform initialization gives
\[
0\le\Phi_0\le H\log A,
\]
and $\Phi_t\ge0$ for every $t$.

Apply Lemma~\ref{lem:npg-three-point-identity} with
\[
u=\pi_t^h(\cdot\mid s),
\qquad
u^+=\pi_{t+1}^h(\cdot\mid s),
\qquad
\nu=\pi^{\star,h}(\cdot\mid s),
\qquad
z=\widehat Q_t^h(s,\cdot),
\]
and $\eta=\eta_t$. This yields
\[
\sum_a
\bigl(\pi^{\star,h}(a\mid s)-\pi_{t+1}^h(a\mid s)\bigr)
\widehat Q_t^h(s,a)
\le
\frac{\kappa_t^h(s)-\kappa_{t+1}^h(s)}{\eta_t}.
\]
Combining this inequality with
Lemma~\ref{lem:optimal-policy-comparison} gives, for $h<H$,
\[
\begin{aligned}
\widetilde V_t^{\star,h}(s)-f_t^h(s)
\le
\frac{\kappa_t^h(s)-\kappa_{t+1}^h(s)}{\eta_t} +
\sum_a\pi^{\star,h}(a\mid s)
\bigl(
\bar P_t^h(\widetilde V_t^{\star,h+1}-f_t^{h+1})
\bigr)(s,a).
\end{aligned}
\]
At $h=H$, the continuation term is absent.

The function $\widetilde V_t^{\star,h+1}-f_t^{h+1}$ is
$\mathcal G_t^{h+1}$-measurable. Fresh sampling and the tower
property therefore imply
\[
\begin{aligned}
\mathbb E\!\left[
\bigl(\bar P_t^h(\widetilde V_t^{\star,h+1}-f_t^{h+1})\bigr)(s,a)
\,\middle|\,
\mathcal F_t
\right] =
\sum_{s'}P^h(s'\mid s,a)
\left(
V^{\star,h+1}(s')
-
\mathbb E[f_t^{h+1}(s')\mid\mathcal F_t]
\right).
\end{aligned}
\]
Taking conditional expectations in the preceding comparison
inequality consequently gives
\[
\begin{aligned}
V^{\star,h}(s)-\mathbb E[f_t^h(s)\mid\mathcal F_t]
& \le
\frac{
\kappa_t^h(s)-\mathbb E[\kappa_{t+1}^h(s)\mid\mathcal F_t]
}{\eta_t}
\\
&\qquad+
\sum_{a,s'}\pi^{\star,h}(a\mid s)P^h(s'\mid s,a)
\left(
V^{\star,h+1}(s')
-
\mathbb E[f_t^{h+1}(s')\mid\mathcal F_t]
\right).
\end{aligned}
\]
Again, omit the continuation term at $H$.

Multiply by $d_\rho^{1\to h,\pi^\star}(s)$, sum over states,
and telescope across horizons. This gives
\[
\eta_t\left(
V^{\star,1}(\rho)
-
\mathbb E[f_t^1(\rho)\mid\mathcal F_t]
\right)
\le
\Phi_t-\mathbb E[\Phi_{t+1}\mid\mathcal F_t].
\]
Taking expectations and summing over $t=0,\ldots,T-1$,
\[
\sum_{t=0}^{T-1}\eta_t
\mathbb E\!\left[
V^{\star,1}(\rho)-f_t^1(\rho)
\right]
\le
\Phi_0-\mathbb E[\Phi_T]
\le H\log A.
\]
\end{proof}

\subsection{Proof of Theorem~\ref{thm:npg-doubling-batches}}
\label{app:proof-npg-doubling-batches}

\begin{proof}
For each iteration,
\[
\begin{aligned}
V^{\star,1}(\rho)-V^{\pi_{t+1},1}(\rho)=
V^{\star,1}(\rho)-f_t^1(\rho) +
f_t^1(\rho)-V^{\pi_{t+1},1}(\rho).
\end{aligned}
\]
Multiply by $\eta_t$, take expectations, and sum over iterations.
Lemmas~\ref{lem:conditional-sampling-error}
and~\ref{lem:npg-weighted-comparison} imply
\[
\begin{aligned}
\sum_{t=0}^{T-1}\eta_t
\mathbb E\!\left[
V^{\star,1}(\rho)-V^{\pi_{t+1},1}(\rho)
\right] \le
H\log A+
6\sum_{t=0}^{T-1}\eta_t
\sqrt{\frac{H^3\log(2SA)}{m_t}}.
\end{aligned}
\]
Every term on the left is nonnegative. Retaining only the
final term and substituting $m_t=2^t$ and $\eta_t=2^t/H$
yields
\[
\begin{aligned}
\mathbb E\!\left[
V^{\star,1}(\rho)-V^{\pi_T,1}(\rho)
\right]
&\le
\frac{H^2\log A}{2^{T-1}}
+
6\sum_{t=0}^{T-1}
2^{t-(T-1)}
\sqrt{\frac{H^3\log(2SA)}{2^t}}
\\
&=
\frac{H^2\log A}{2^{T-1}}
+
6\sqrt{\frac{H^3\log(2SA)}{2^{T-1}}}
\sum_{t=0}^{T-1}2^{-(T-1-t)/2}
\\
&\le
\frac{H^2\log A}{2^{T-1}}
+
(12+6\sqrt2)
\sqrt{\frac{H^3\log(2SA)}{2^{T-1}}},
\end{aligned}
\]
where the last inequality uses
\[
\sum_{t=0}^{T-1}2^{-(T-1-t)/2}
\le
\frac{1}{1-2^{-1/2}}
=
2+\sqrt2.
\]

If $2^{T-1}\ge16H\log(2SA)$, then
\[
\begin{aligned}
\frac{H^2\log A}{2^{T-1}}
&\le
\sqrt{\frac{H^3\log(2SA)}{2^{T-1}}}
\sqrt{\frac{H\log(2SA)}{2^{T-1}}}
\\
&\le
\frac14\sqrt{\frac{H^3\log(2SA)}{2^{T-1}}}.
\end{aligned}
\]
Since $12+\tfrac14+6\sqrt2<21$, it follows that
\[
\mathbb E\!\left[
V^{\star,1}(\rho)-V^{\pi_T,1}(\rho)
\right]
\le
21\sqrt{\frac{H^3\log(2SA)}{2^{T-1}}}.
\]

If $2^{T-1}<16H\log(2SA)$, then
\[
21\sqrt{\frac{H^3\log(2SA)}{2^{T-1}}}
>
\frac{21H}{4}
>
H,
\]
and the value range gives the required bound. Therefore,
in both cases,
\[
\mathbb E\!\left[
V^{\star,1}(\rho)-V^{\pi_T,1}(\rho)
\right]
\le
\min\!\left\{
H,\;
21\sqrt{\frac{H^3\log(2SA)}{2^{T-1}}}
\right\}.
\]
\end{proof}

\subsection{Additional lemmas for NPG with constant step size}
\label{app:npg-constant-step-lemmas}

Fix an iteration $t$, an integer batch size $m\ge2$, and step size $0<\eta\le1/(4H^2)$. The original update starts from
$\pi_t$ and produces $\pi_{t+1}$. Throughout this subsection,
write
\[
\mathbb E_t[\cdot]:=\mathbb E[\cdot\mid\mathcal F_t],
\qquad
\zeta:=\eta H.
\]
Thus $H\zeta\le1/4$ and $\zeta\le1/8$.

For each horizon $j<H$, construct an auxiliary iteration
using the same initial policy and oracle samples, except that
the sampled continuation at horizon $j$ is replaced by its
exact expectation. For $h>j$, set
\[
\widetilde Q_{t,j}^{\,h}=\widehat Q_t^h,
\qquad
\widetilde\pi_{t+1,j}^{\,h}=\pi_{t+1}^h,
\qquad
\widetilde f_{t,j}^{\,h}=f_t^h.
\]
At horizon $j$, set
\[
\widetilde Q_{t,j}^{\,j}(s,a)
=
R^j(s,a)+(P^j f_t^{j+1})(s,a).
\]
For $h=j-1,\ldots,1$, compute
\[
\widetilde Q_{t,j}^{\,h}(s,a)
=
R^h(s,a)
+
(\bar P_t^h\widetilde f_{t,j}^{\,h+1})(s,a).
\]
At each $h\le j$, apply \eqref{eq:sample-based-npg-update}
with base distribution $\pi_t^h(\cdot\mid s)$, step size
$\eta$, and action values $\widetilde Q_{t,j}^{\,h}$ to obtain
$\widetilde\pi_{t+1,j}^{\,h}$. Then set
\[
\widetilde f_{t,j}^{\,h}(s)
=
\sum_a
\widetilde\pi_{t+1,j}^{\,h}(a\mid s)
\widetilde Q_{t,j}^{\,h}(s,a).
\]
The range bounds of Lemma~\ref{lem:sampled-value-bounds}
also hold for these auxiliary quantities.

Define
\[
\widetilde w_{t,j}^{\,j}(s,a)
:=
d_\rho^{1\to j,\widetilde\pi_{t+1,j}}(s)
\widetilde\pi_{t+1,j}^{\,j}(a\mid s).
\]
These auxiliary quantities are used only in the analysis.
The distributions $w_t^j$ and $\widetilde w_{t,j}^{\,j}$
have the same support because both policies have full support
and use the same initial distribution and transition kernels.
For their KL divergence, sum over state-action pairs in this
common support.

We also use the visitation probabilities under the policy
before the update:
\[
d_{t,0}^h(s):=d_\rho^{1\to h,\pi_t}(s),
\qquad
w_{t,0}^h(s,a):=d_{t,0}^h(s)\pi_t^h(a\mid s).
\]
These quantities are $\mathcal F_t$-measurable. Define
\[
\mathcal V_{t,0}^h
:=
\sum_{s,a}w_{t,0}^h(s,a)v_t^h(s,a),
\qquad
W_{t,0}:=\sum_{h=1}^{H}\mathcal V_{t,0}^h.
\]

\begin{lem}
\label{lem:conditional-signed-sampling-error}
For every constant $\lambda\in(0,m/H]$,
\[
\begin{aligned}
\mathbb E_t\!\left[
f_t^1(\rho)-V^{\pi_{t+1},1}(\rho)
\right]
\le
\frac1\lambda
\sum_{j=1}^{H-1}
\mathbb E_t\!\left[
D_{\mathrm{KL}}\!\left(
w_t^j\middle\|\widetilde w_{t,j}^{\,j}
\right)
\right]
+
\frac{\lambda}{m}\mathbb E_t[W_t]
\end{aligned}
\]
almost surely.
\end{lem}

\begin{proof}
Fix $j<H$. Let $\mathcal H_t^j$ be the $\sigma$-field
generated by $\mathcal F_t$ and all samples from iteration
$t$ outside horizon $j$. The quantities $f_t^{j+1}$,
$v_t^j$, and the entire auxiliary policy
$\widetilde\pi_{t+1,j}$ are $\mathcal H_t^j$-measurable.
Conditional on $\mathcal H_t^j$, the samples at horizon $j$
retain their independent oracle distributions.

The exponential-moment argument in
Lemma~\ref{lem:weighted-sampling-error} gives
\[
\mathbb E\!\left[
\exp\!\left(
\lambda e_t^j(s,a)-\frac{\lambda^2}{m}v_t^j(s,a)
\right)
\,\middle|\,
\mathcal H_t^j
\right]
\le1.
\]
Set
\[
\psi_t^j(s,a)
:=
\lambda e_t^j(s,a)-\frac{\lambda^2}{m}v_t^j(s,a).
\]
For probability distributions $p,q$ on a finite common
support and a real-valued function $\psi$,
\[
\sum_x p(x)\psi(x)
\le
D_{\mathrm{KL}}(p\|q)
+
\log\!\left(\sum_x q(x)e^{\psi(x)}\right).
\]
Indeed, this follows by expanding the nonnegative KL
divergence from $p$ to the distribution proportional to
$q(x)e^{\psi(x)}$.

Apply this inequality with $p=w_t^j$,
$q=\widetilde w_{t,j}^{\,j}$, and $\psi=\psi_t^j$.
Conditional Jensen inequality and the measurability of
$\widetilde w_{t,j}^{\,j}$ give
\[
\begin{aligned}
\mathbb E\!\left[
\log\!\left(
\sum_{s,a}\widetilde w_{t,j}^{\,j}(s,a)
e^{\psi_t^j(s,a)}
\right)
\,\middle|\,
\mathcal H_t^j
\right]
\le
\log\!\left(
\sum_{s,a}\widetilde w_{t,j}^{\,j}(s,a)
\mathbb E[e^{\psi_t^j(s,a)}\mid\mathcal H_t^j]
\right)
\le0.
\end{aligned}
\]
Taking conditional expectations given $\mathcal F_t$
therefore yields
\[
\begin{aligned}
\lambda\mathbb E_t\!\left[
\sum_{s,a}w_t^j(s,a)e_t^j(s,a)
\right]
\le
\mathbb E_t\!\left[
D_{\mathrm{KL}}\!\left(
w_t^j\middle\|\widetilde w_{t,j}^{\,j}
\right)
\right] +
\frac{\lambda^2}{m}
\mathbb E_t\!\left[
\sum_{s,a}w_t^j(s,a)v_t^j(s,a)
\right].
\end{aligned}
\]
All terms are integrable. In particular, the range bounds
and the normalized updates imply
\[
e^{-2\eta H}
\le
\frac{\pi_{t+1}^h(a\mid s)}
     {\widetilde\pi_{t+1,j}^{\,h}(a\mid s)}
\le
e^{2\eta H}.
\]
Comparing trajectory probabilities through horizon $j$
then gives
\[
D_{\mathrm{KL}}\!\left(
w_t^j\middle\|\widetilde w_{t,j}^{\,j}
\right)
\le2\eta Hj.
\]
Finally, sum over $j=1,\ldots,H-1$, divide by $\lambda$,
and apply Lemma~\ref{lem:sampled-value-error-identity}.
\end{proof}

\begin{lem}
\label{lem:visitation-stability}
For every integer $m\ge2$ and step size
$0<\eta\le1/(4H^2)$,
\[
\sum_{j=1}^{H-1}
\mathbb E_t\!\left[
D_{\mathrm{KL}}\!\left(
w_t^j\middle\|\widetilde w_{t,j}^{\,j}
\right)
\right]
\le
\frac{14\eta^2}{m}\mathbb E_t[W_t]
\qquad\text{almost surely}.
\]
\end{lem}

\begin{proof}
The normalized update and the action-value range imply
\[
e^{-\zeta}
\le
\frac{\pi_{t+1}^h(a\mid s)}{\pi_t^h(a\mid s)}
\le
e^\zeta.
\]
Comparing trajectory probabilities gives
\[
d_\rho^{1\to h,\pi_{t+1}}(s)
\le
e^{(h-1)\zeta}d_{t,0}^h(s),
\qquad
e^{-h\zeta}w_{t,0}^h(s,a)
\le
w_t^h(s,a)
\le
e^{h\zeta}w_{t,0}^h(s,a).
\tag{*}
\]

Fix $j<H$. Within this proof, suppress the fixed indices
$t,j$ in the following differences:
\[
\omega^h(s,a)
:=
\widehat Q_t^h(s,a)-\widetilde Q_{t,j}^{\,h}(s,a),
\qquad
\xi^h(s)
:=
f_t^h(s)-\widetilde f_{t,j}^{\,h}(s).
\]
Then
\[
\omega^j=e_t^j,
\qquad
\omega^h=\bar P_t^h\xi^{h+1}
\quad(h<j).
\]
Define
\[
\Omega_h
:=
\sum_s d_{t,0}^h(s)
\operatorname{Var}_{a\sim\pi_t^h(\cdot\mid s)}
\bigl(\omega^h(s,a)\bigr),
\qquad
\Xi_h
:=
\sum_s d_{t,0}^h(s)\bigl(\xi^h(s)\bigr)^2.
\]

At a fixed state and horizon, write
$\pi=\pi_{t+1}^h(\cdot\mid s)$,
$\widetilde\pi=\widetilde\pi_{t+1,j}^{\,h}(\cdot\mid s)$,
$p=\pi_t^h(\cdot\mid s)$, and $\omega=\omega^h(s,\cdot)$.
The two updates satisfy
$\pi(a)\propto\widetilde\pi(a)e^{\eta\omega(a)}$, so
\[
D_{\mathrm{KL}}(\pi\|\widetilde\pi)
=
\log\mathbb E_{a\sim\pi}
\exp\!\left(
-\eta\bigl(\omega(a)-\mathbb E_{a'\sim\pi}\omega(a')\bigr)
\right).
\]
The centered exponent has magnitude at most $2\zeta\le1/4$.
Using $e^x\le1+x+x^2$ and $\log(1+x)\le x$, we obtain
\[
D_{\mathrm{KL}}(\pi\|\widetilde\pi)
\le
\eta^2\operatorname{Var}_{\pi}(\omega)
\le
\eta^2e^\zeta\operatorname{Var}_{p}(\omega).
\]
The final inequality follows by centering at the $p$-mean
and using $\pi\le e^\zeta p$.

By the chain rule for trajectory KL divergence and
marginalization to $(S^j,A^j)$,
\[
\begin{aligned}
D_{\mathrm{KL}}\!\left(
w_t^j\middle\|\widetilde w_{t,j}^{\,j}
\right)
&\le
\sum_{h=1}^{j}\sum_s
d_\rho^{1\to h,\pi_{t+1}}(s)
D_{\mathrm{KL}}\!\left(
\pi_{t+1}^h(\cdot\mid s)
\middle\|
\widetilde\pi_{t+1,j}^{\,h}(\cdot\mid s)
\right)
\\
&\le
\eta^2e^{H\zeta}\sum_{h=1}^{j}\Omega_h.
\end{aligned}
\]

For $h<j$, define
\[
\mu^h(s)
:=
\sum_a\pi_t^h(a\mid s)(P^h\xi^{h+1})(s,a),
\qquad
\epsilon^h
:=
(\bar P_t^h-P^h)\xi^{h+1},
\]
and
\[
\begin{aligned}
\Xi_h^-
&:=
\sum_s d_{t,0}^h(s)\bigl(\mu^h(s)\bigr)^2,
\\
\Lambda_h
&:=
\sum_s d_{t,0}^h(s)
\operatorname{Var}_{a\sim\pi_t^h(\cdot\mid s)}
\bigl((P^h\xi^{h+1})(s,a)\bigr),
\\
\Theta_h
&:=
\sum_{s,a}w_{t,0}^h(s,a)
\operatorname{Var}_{Y\sim P^h(\cdot\mid s,a)}
\bigl(\xi^{h+1}(Y)\bigr).
\end{aligned}
\]
The law of total variance and the visitation recursion
under $\pi_t$ give the identity
\[
\Xi_{h+1}=\Xi_h^-+\Lambda_h+\Theta_h.
\]

Conditional on $\mathcal G_t^{h+1}$, both continuation
estimates at horizon $h+1$ are fixed. Hence each
$\epsilon^h(s,a)$ has conditional mean zero and conditional
variance
\[
\frac1m
\operatorname{Var}_{Y\sim P^h(\cdot\mid s,a)}
\bigl(\xi^{h+1}(Y)\bigr).
\]
Expanding the action variance of
$\omega^h=P^h\xi^{h+1}+\epsilon^h$ therefore gives
\[
\mathbb E[\Omega_h\mid\mathcal G_t^{h+1}]
\le
\Lambda_h+\frac{\Theta_h}{m}.
\]

Decompose
\[
\xi^h(s)
=
\mu^h(s)+r_2^h(s)+r_1^h(s)+r_3^h(s),
\]
where
\[
\begin{aligned}
r_2^h(s)
&:=
\sum_a\pi_t^h(a\mid s)\epsilon^h(s,a),
\\
r_1^h(s)
&:=
\sum_a
\bigl(\pi_{t+1}^h(a\mid s)-\pi_t^h(a\mid s)\bigr)
\omega^h(s,a),
\\
r_3^h(s)
&:=
\sum_a
\bigl(\pi_{t+1}^h(a\mid s)
-\widetilde\pi_{t+1,j}^{\,h}(a\mid s)\bigr)
\widetilde Q_{t,j}^{\,h}(s,a).
\end{aligned}
\]
The conditional mean of $r_2^h(s)$ is zero, and Jensen's
inequality gives
\[
\mathbb E\!\left[
\sum_s d_{t,0}^h(s)\bigl(r_2^h(s)\bigr)^2
\,\middle|\,
\mathcal G_t^{h+1}
\right]
\le\frac{\Theta_h}{m}.
\]
Centering $\omega^h(s,\cdot)$ at its $\pi_t^h$-mean yields
\[
|r_1^h(s)|
\le
(e^\zeta-1)
\sqrt{
\operatorname{Var}_{a\sim\pi_t^h(\cdot\mid s)}
\bigl(\omega^h(s,a)\bigr)
}.
\]
Since $\widetilde Q_{t,j}^{\,h}(s,a)\in[0,H]$,
Pinsker's inequality and the preceding KL bound imply
\[
|r_3^h(s)|
\le
\frac{\zeta e^{\zeta/2}}{\sqrt2}
\sqrt{
\operatorname{Var}_{a\sim\pi_t^h(\cdot\mid s)}
\bigl(\omega^h(s,a)\bigr)
}.
\]
For $\zeta\le1/8$,
\[
e^\zeta-1\le\frac87\zeta,
\qquad
\left(\frac87+\frac{e^{1/16}}{\sqrt2}\right)^2<4.
\]
Consequently,
\[
\sum_s d_{t,0}^h(s)
\bigl(r_1^h(s)+r_3^h(s)\bigr)^2
\le4\zeta^2\Omega_h.
\]

Applying
$(a+b)^2\le(1+\zeta)a^2+(1+1/\zeta)b^2$
to $a=\mu^h+r_2^h$ and $b=r_1^h+r_3^h$, and using
$4\zeta(1+\zeta)\le4.5\zeta$, gives
\[
\mathbb E[\Xi_h\mid\mathcal G_t^{h+1}]
\le
(1+\zeta)\left(\Xi_h^-+\frac{\Theta_h}{m}\right)
+
4.5\zeta\left(\Lambda_h+\frac{\Theta_h}{m}\right).
\]
Substitute $\Xi_h^-=\Xi_{h+1}-\Lambda_h-\Theta_h$.
Since
\[
1-3.5\zeta\ge\frac9{16},
\qquad
1+\zeta-\frac{1+5.5\zeta}{m}\ge\frac9{32},
\]
taking conditional expectations given $\mathcal F_t$ yields
\[
\frac12\mathbb E_t[\Lambda_h]
+
\frac14\mathbb E_t[\Theta_h]
\le
(1+\zeta)\mathbb E_t[\Xi_{h+1}]
-
\mathbb E_t[\Xi_h].
\]
In particular,
$\mathbb E_t[\Xi_h]\le e^{H\zeta}\mathbb E_t[\Xi_j]$
for $h\le j$. Summing the preceding inequality over $h<j$
therefore gives
\[
\sum_{h<j}\mathbb E_t[\Lambda_h+\Theta_h]
\le
4(1+H\zeta e^{H\zeta})\mathbb E_t[\Xi_j]
\le
5.3\,\mathbb E_t[\Xi_j].
\]
For $j=1$, the sum is empty.

At horizon $j$,
\[
\xi^j(s)
=
\sum_a\pi_{t+1}^j(a\mid s)e_t^j(s,a)
+
r_3^j(s).
\]
Cauchy--Schwarz and
$\pi_{t+1}^j\le e^\zeta\pi_t^j$, together with the bound
on $r_3^j$, imply
\[
\begin{aligned}
\bigl(\xi^j(s)\bigr)^2
&\le
e^\zeta
\left(1+\frac{\zeta}{\sqrt2}\right)^2
\sum_a\pi_t^j(a\mid s)\bigl(e_t^j(s,a)\bigr)^2
\\
&\le
1.35\sum_a\pi_t^j(a\mid s)\bigl(e_t^j(s,a)\bigr)^2.
\end{aligned}
\]
Conditional on $\mathcal G_t^{j+1}$,
\[
\mathbb E[(e_t^j(s,a))^2\mid\mathcal G_t^{j+1}]
=
\frac{v_t^j(s,a)}{m}.
\]
Thus
\[
\mathbb E_t[\Xi_j]
\le
\frac{1.35}{m}\mathbb E_t[\mathcal V_{t,0}^j],
\qquad
\mathbb E_t[\Omega_j]
\le
\frac1m\mathbb E_t[\mathcal V_{t,0}^j].
\]
Combining these estimates gives
\[
\sum_{h=1}^{j}\mathbb E_t[\Omega_h]
\le
\frac{1+5.3\cdot1.35}{m}
\mathbb E_t[\mathcal V_{t,0}^j].
\]
Using $(*)$ to replace the baseline visitation weights,
we conclude that
\[
\begin{aligned}
\mathbb E_t\!\left[
D_{\mathrm{KL}}\!\left(
w_t^j\middle\|\widetilde w_{t,j}^{\,j}
\right)
\right]
& \le
\frac{\eta^2e^{2H\zeta}(1+5.3\cdot1.35)}{m}
\mathbb E_t\!\left[
\sum_{s,a}w_t^j(s,a)v_t^j(s,a)
\right]
\\
&\quad\le
\frac{14\eta^2}{m}
\mathbb E_t\!\left[
\sum_{s,a}w_t^j(s,a)v_t^j(s,a)
\right],
\end{aligned}
\]
because $e^{2H\zeta}\le e^{1/2}$ and
$e^{1/2}(1+5.3\cdot1.35)<14$.
Summing over $j=1,\ldots,H-1$ proves the lemma.
\end{proof}

\begin{lem}
\label{lem:small-step-variance-bias}
For every integer $m\ge2$ and step size
$0<\eta\le1/(4H^2)$,
\[
\mathbb E_t[W_t]\le4H^2
\]
and
\[
\mathbb E_t\!\left[
f_t^1(\rho)-V^{\pi_{t+1},1}(\rho)
\right]
\le
\frac{30\eta H^2}{m}
\]
almost surely.
\end{lem}

\begin{proof}
Define
\[
U_h
:=
\sum_s d_{t,0}^h(s)\bigl(f_t^h(s)\bigr)^2,
\qquad h\in[H],
\]
and set $U_{H+1}=0$. Jensen's inequality and
$\pi_{t+1}^h\le e^\zeta\pi_t^h$ give
\[
e^{-\zeta}U_h
\le
\sum_{s,a}w_{t,0}^h(s,a)
\bigl(\widehat Q_t^h(s,a)\bigr)^2.
\]
For $h<H$, conditional on $\mathcal G_t^{h+1}$,
\[
\mathbb E\!\left[
\bigl(\widehat Q_t^h(s,a)\bigr)^2
\,\middle|\,
\mathcal G_t^{h+1}
\right]
=
\bigl(R^h(s,a)+(P^h f_t^{h+1})(s,a)\bigr)^2
+
\frac{v_t^h(s,a)}{m}.
\]
The reward and continuation bounds imply
\[
\bigl(R^h(s,a)+(P^h f_t^{h+1})(s,a)\bigr)^2
\le
\bigl((P^h f_t^{h+1})(s,a)\bigr)^2
+
2(H-h)+1.
\]
Furthermore, the visitation recursion under $\pi_t$ gives
\[
\sum_{s,a}w_{t,0}^h(s,a)
\bigl((P^h f_t^{h+1})(s,a)\bigr)^2
=
U_{h+1}-\mathcal V_{t,0}^h.
\]
Hence
\[
e^{-\zeta}\mathbb E_t[U_h]
\le
\mathbb E_t[U_{h+1}]
-
\left(1-\frac1m\right)
\mathbb E_t[\mathcal V_{t,0}^h]
+
2(H-h)+1.
\]
At $h=H$, the same inequality holds because
$\widehat Q_t^H=R^H$, $U_{H+1}=0$, and
$\mathcal V_{t,0}^H=0$.

Sum over $h=1,\ldots,H$, use
$\sum_{h=1}^{H}[2(H-h)+1]=H^2$, and discard
$-\mathbb E_t[U_1]$. This yields
\[
\left(1-\frac1m\right)\mathbb E_t[W_{t,0}]
\le
H^2+(1-e^{-\zeta})\sum_{h=1}^{H}\mathbb E_t[U_h].
\]
Since $U_h\le H^2$ and $H\zeta\le1/4$,
\[
\left(1-\frac1m\right)\mathbb E_t[W_{t,0}]
\le
H^2+\zeta H^3
\le
\frac54H^2.
\]
Using $m\ge2$ gives
\[
\mathbb E_t[W_{t,0}]\le\frac52H^2.
\]
The visitation-ratio bound in the proof of
Lemma~\ref{lem:visitation-stability} implies
$W_t\le e^{H\zeta}W_{t,0}$. Therefore,
\[
\mathbb E_t[W_t]
\le
\frac52e^{1/4}H^2
<
4H^2.
\]

For the bias bound, choose $\lambda=\sqrt{14}\eta$.
This choice is admissible in
Lemma~\ref{lem:conditional-signed-sampling-error}, since
\[
\lambda H=\sqrt{14}\zeta
\le\frac{\sqrt{14}}8<m.
\]
Combining that lemma with
Lemma~\ref{lem:visitation-stability} gives
\[
\begin{aligned}
\mathbb E_t\!\left[
f_t^1(\rho)-V^{\pi_{t+1},1}(\rho)
\right]
&\le
\left(
\frac{14\eta^2}{\lambda m}+\frac{\lambda}{m}
\right)\mathbb E_t[W_t]
\\
&=
\frac{2\sqrt{14}\eta}{m}\mathbb E_t[W_t]
\\
&\le
\frac{8\sqrt{14}\eta H^2}{m}
<
\frac{30\eta H^2}{m}.
\end{aligned}
\]
\end{proof}

For a Markov policy $\nu$, define
\[
\Phi_t(\nu)
:=
\sum_{h=1}^{H}\sum_s
d_\rho^{1\to h,\nu}(s)
D_{\mathrm{KL}}\!\left(
\nu^h(\cdot\mid s)
\middle\|
\pi_t^h(\cdot\mid s)
\right).
\]

\begin{lem}[Comparison with a policy determined before sampling]
\label{lem:history-measurable-policy-comparison}
Let $\nu$ be an $\mathcal F_t$-measurable Markov policy.
Then
\[
\eta\left(
V^{\nu,1}(\rho)-\mathbb E_t[f_t^1(\rho)]
\right)
\le
\Phi_t(\nu)-\mathbb E_t[\Phi_{t+1}(\nu)]
\qquad\text{almost surely}.
\]
\end{lem}

\begin{proof}
Conditional on $\mathcal F_t$, the policy $\nu$ and its
visitation probabilities are fixed. Define an auxiliary
evaluation of $\nu$ using the samples from iteration $t$:
\[
\widetilde V_t^{\nu,H}(s)
:=
\sum_a\nu^H(a\mid s)R^H(s,a),
\]
and, for $h<H$,
\[
\widetilde V_t^{\nu,h}(s)
:=
\sum_a\nu^h(a\mid s)
\left[
R^h(s,a)
+
(\bar P_t^h\widetilde V_t^{\nu,h+1})(s,a)
\right].
\]
The argument of Lemma~\ref{lem:optimal-policy-comparison},
applied conditionally on $\mathcal F_t$, gives
\[
\mathbb E_t[\widetilde V_t^{\nu,h}(s)]
=
V^{\nu,h}(s).
\]
It also gives, for $h<H$,
\[
\begin{aligned}
\widetilde V_t^{\nu,h}(s)-f_t^h(s)
=
\sum_a
\bigl(\nu^h(a\mid s)-\pi_{t+1}^h(a\mid s)\bigr)
\widehat Q_t^h(s,a) +
\sum_a\nu^h(a\mid s)
\bigl(
\bar P_t^h(\widetilde V_t^{\nu,h+1}-f_t^{h+1})
\bigr)(s,a),
\end{aligned}
\]
with the continuation term omitted at $H$.

Apply Lemma~\ref{lem:npg-three-point-identity} to the first
sum. Taking conditional expectations given $\mathcal F_t$
and using fresh sampling in the continuation term yields
\[
\begin{aligned}
V^{\nu,h}(s)-\mathbb E_t[f_t^h(s)]
&\le
\frac1\eta
\left[
D_{\mathrm{KL}}\!\left(
\nu^h(\cdot\mid s)\middle\|\pi_t^h(\cdot\mid s)
\right)
-
\mathbb E_t\!\left[
D_{\mathrm{KL}}\!\left(
\nu^h(\cdot\mid s)\middle\|\pi_{t+1}^h(\cdot\mid s)
\right)
\right]
\right]
\\
&\qquad \qquad+
\sum_{a,s'}\nu^h(a\mid s)P^h(s'\mid s,a)
\left[
V^{\nu,h+1}(s')-\mathbb E_t[f_t^{h+1}(s')]
\right].
\end{aligned}
\]
At $H$, the last term is absent.

Multiply by $d_\rho^{1\to h,\nu}(s)$, sum over states,
and telescope across horizons using the visitation recursion
for $\nu$. The resulting inequality is
\[
\eta\left(
V^{\nu,1}(\rho)-\mathbb E_t[f_t^1(\rho)]
\right)
\le
\Phi_t(\nu)-\mathbb E_t[\Phi_{t+1}(\nu)].
\]
The conditional expectations are well-defined: full support
and the finite bound on $-\log\pi_t^h(a\mid s)$ established
in Lemma~\ref{lem:npg-weighted-comparison} ensure integrability
of the KL terms, including when $\nu$ is random.
\end{proof}

\subsection{Proof of Theorem~\ref{thm:npg-constant-step}}
\label{app:proof-npg-constant-step}

\begin{proof}
For $t=1,\ldots,T$, define
\[
\delta_t
:=
\mathbb E\!\left[
V^{\star,1}(\rho)-V^{\pi_t,1}(\rho)
\right],
\qquad
b:=\frac{30\eta H^2}{m}.
\]
Combining Lemmas~\ref{lem:npg-weighted-comparison}
and~\ref{lem:small-step-variance-bias}, and dividing by
$\eta T$, gives
\[
\frac1T\sum_{t=1}^{T}\delta_t
\le
\frac{H\log A}{\eta T}+b.
\]

Fix $t\in\{1,\ldots,T-1\}$. For each
$t'\in\{t,\ldots,T-1\}$, the policy $\pi_t$ is
$\mathcal F_{t'}$-measurable.
Apply Lemma~\ref{lem:history-measurable-policy-comparison}
at iteration $t'$ with $\nu=\pi_t$, and use
Lemma~\ref{lem:small-step-variance-bias} to obtain
\[
\begin{aligned}
\eta\left(
V^{\pi_t,1}(\rho)
-
\mathbb E[V^{\pi_{t'+1},1}(\rho)\mid\mathcal F_{t'}]
\right) \le
\Phi_{t'}(\pi_t)
-
\mathbb E[\Phi_{t'+1}(\pi_t)\mid\mathcal F_{t'}]
+\eta b.
\end{aligned}
\]
Taking expectations and summing over $t'=t,\ldots,T-1$,
the potential terms telescope. Since
$\Phi_t(\pi_t)=0$ and $\Phi_T(\pi_t)\ge0$,
\[
\sum_{t'=t+1}^{T}
\mathbb E\!\left[
V^{\pi_t,1}(\rho)-V^{\pi_{t'},1}(\rho)
\right]
\le
(T-t)b.
\]
Equivalently,
\[
\sum_{t'=t+1}^{T}(\delta_{t'}-\delta_t)
\le
(T-t)b.
\]

For $t=1,\ldots,T$, define
\[
\overline\delta_t
:=
\frac{1}{T-t+1}\sum_{t'=t}^{T}\delta_{t'}.
\]
For $t<T$, the preceding inequality implies
$\overline\delta_{t+1}-\delta_t\le b$. Hence
\[
\overline\delta_{t+1}-\overline\delta_t
=
\frac{\overline\delta_{t+1}-\delta_t}{T-t+1}
\le
\frac{b}{T-t+1}.
\]
Summing over $t=1,\ldots,T-1$ and using
$\overline\delta_T=\delta_T$ gives
\[
\begin{aligned}
\delta_T
&\le
\overline\delta_1
+
b\sum_{t=1}^{T-1}\frac{1}{T-t+1}
\\
&\le
\frac{H\log A}{\eta T}
+
b\sum_{t=1}^{T}\frac1t.
\end{aligned}
\]
For $T=1$, the same conclusion follows directly from the
initial average bound. Finally, $\delta_T\le H$ by the value
range. Therefore,
\[
\mathbb E\!\left[
V^{\star,1}(\rho)-V^{\pi_T,1}(\rho)
\right]
\le
\min\!\left\{
H,\;
\frac{H\log A}{\eta T}
+
\frac{30\eta H^2}{m}\sum_{t=1}^{T}\frac1t
\right\}.
\]
\end{proof}

\subsection{Supporting lemma for linearly decreasing step size}
\label{app:npg-linear-step-lemma}

Lemmas~\ref{lem:small-step-variance-bias}
and~\ref{lem:history-measurable-policy-comparison} apply
separately at each iteration with $\eta=\eta_t$.
We first establish a bound for deterministic step size schedules,
then specialize it to the linearly decreasing schedule.

\begin{lem}[Last-policy bound for deterministic step size schedules]
\label{lem:npg-varying-step-last-policy}
Fix an integer $T\ge1$. Consider the sample-based NPG procedure
with uniform initialization, $m_t=2$, and a deterministic
sequence $(\eta_t)_{t=0}^{T-1}$ satisfying
\[
0<\eta_t\le\frac1{4H^2},
\qquad t=0,\ldots,T-1.
\]
No monotonicity assumption is imposed on this sequence.
Define
\[
\Gamma_t
:=
\sum_{t'=t}^{T-1}\eta_{t'},
\qquad t=0,\ldots,T-1.
\]
Then
\[
\mathbb E\!\left[
V^{\star,1}(\rho)-V^{\pi_T,1}(\rho)
\right]
\le
\frac{H\log A}{\Gamma_0}
+
15H^2\sum_{t=0}^{T-1}\frac{\eta_t^2}{\Gamma_t}.
\]
\end{lem}

\begin{proof}
Retain the notation
$\delta_t=\mathbb E[V^{\star,1}(\rho)-V^{\pi_t,1}(\rho)]$
from the proof of Theorem~\ref{thm:npg-constant-step}.
At iteration $t$, Lemma~\ref{lem:small-step-variance-bias}
with $m=2$ and $\eta=\eta_t$ gives
\[
\mathbb E\!\left[
f_t^1(\rho)-V^{\pi_{t+1},1}(\rho)
\,\middle|\,
\mathcal F_t
\right]
\le15H^2\eta_t.
\]
Combining this with Lemma~\ref{lem:npg-weighted-comparison}
yields
\[
\sum_{t=0}^{T-1}\eta_t\delta_{t+1}
\le
H\log A+15H^2\sum_{t=0}^{T-1}\eta_t^2.
\]

For $T\ge2$, fix $t\in\{0,\ldots,T-2\}$.
The policy $\pi_{t+1}$ is $\mathcal F_{t'}$-measurable for
every $t'\ge t+1$. Applying
Lemma~\ref{lem:history-measurable-policy-comparison}
at iteration $t'$ with comparator $\pi_{t+1}$, and adding
the conditional sampling-bias bound, gives
\[
\begin{aligned}
\eta_{t'}\left(
V^{\pi_{t+1},1}(\rho)
-
\mathbb E[V^{\pi_{t'+1},1}(\rho)\mid\mathcal F_{t'}]
\right) \le
\Phi_{t'}(\pi_{t+1})
-
\mathbb E[\Phi_{t'+1}(\pi_{t+1})\mid\mathcal F_{t'}]
+
15H^2\eta_{t'}^2.
\end{aligned}
\]
Take expectations and sum over $t'=t+1,\ldots,T-1$.
Since
\[
\Phi_{t+1}(\pi_{t+1})=0,
\qquad
\Phi_T(\pi_{t+1})\ge0,
\]
the potential terms telescope to give
\[
\sum_{t'=t+1}^{T-1}
\eta_{t'}(\delta_{t'+1}-\delta_{t+1})
\le
15H^2\sum_{t'=t+1}^{T-1}\eta_{t'}^2.
\]

Define the weighted suffix averages
\[
\overline\delta_t^{\,\eta}
:=
\frac1{\Gamma_t}
\sum_{t'=t}^{T-1}\eta_{t'}\delta_{t'+1},
\qquad t=0,\ldots,T-1.
\]
The initial weighted bound gives
\[
\overline\delta_0^{\,\eta}
\le
\frac{H\log A}{\Gamma_0}
+
\frac{15H^2}{\Gamma_0}
\sum_{t=0}^{T-1}\eta_t^2.
\]
For $t<T-1$, the suffix comparison implies
\[
\begin{aligned}
\overline\delta_{t+1}^{\,\eta}
-
\overline\delta_t^{\,\eta}
&=
\frac{\eta_t}{\Gamma_t}
\left(
\overline\delta_{t+1}^{\,\eta}-\delta_{t+1}
\right)
\\
&\le
15H^2
\frac{\eta_t}{\Gamma_t\Gamma_{t+1}}
\sum_{t'=t+1}^{T-1}\eta_{t'}^2.
\end{aligned}
\]
Because $\Gamma_t-\Gamma_{t+1}=\eta_t$,
\[
\frac{\eta_t}{\Gamma_t\Gamma_{t+1}}
=
\frac1{\Gamma_{t+1}}-\frac1{\Gamma_t}.
\]
Consequently, for each $t'\in\{0,\ldots,T-1\}$,
\[
\frac1{\Gamma_0}
+
\sum_{t=0}^{t'-1}
\frac{\eta_t}{\Gamma_t\Gamma_{t+1}}
=
\frac1{\Gamma_{t'}},
\]
where the sum is empty when $t'=0$.
Summing the successive differences and using
$\overline\delta_{T-1}^{\,\eta}=\delta_T$ therefore yields
\[
\delta_T
\le
\frac{H\log A}{\Gamma_0}
+
15H^2\sum_{t=0}^{T-1}\frac{\eta_t^2}{\Gamma_t}.
\]
For $T=1$, this conclusion follows directly from the initial
weighted bound.
\end{proof}

\subsection{Proof of Theorem~\ref{thm:npg-linear-step}}
\label{app:proof-npg-linear-step}

\begin{proof}
Under the prescribed schedule,
\[
\eta_t=\eta_{\max}\frac{T-t}{T},
\qquad t=0,\ldots,T-1,
\]
every step satisfies $0<\eta_t\le1/(4H^2)$. Moreover,
\[
\Gamma_t
=
\frac{\eta_{\max}}{T}
\sum_{t'=t}^{T-1}(T-t')
=
\frac{\eta_{\max}(T-t)(T-t+1)}{2T}.
\]
Hence
\[
\Gamma_0=\frac{\eta_{\max}(T+1)}2,
\qquad
\frac{\eta_t^2}{\Gamma_t}
=
\frac{2\eta_{\max}}{T}
\frac{T-t}{T-t+1}
\le
\frac{2\eta_{\max}}{T}.
\]
Applying Lemma~\ref{lem:npg-varying-step-last-policy} gives
\[
\begin{aligned}
\mathbb E\!\left[
V^{\star,1}(\rho)-V^{\pi_T,1}(\rho)
\right]
&\le
\frac{2H\log A}{\eta_{\max}(T+1)}
+
15H^2\sum_{t=0}^{T-1}\frac{\eta_t^2}{\Gamma_t}
\\
&\le
\frac{2H\log A}{\eta_{\max}(T+1)}
+
30\eta_{\max}H^2.
\end{aligned}
\]
Combining this inequality with the value-range bound proves
\[
\mathbb E\!\left[
V^{\star,1}(\rho)-V^{\pi_T,1}(\rho)
\right]
\le
\min\!\left\{
H,\;
\frac{2H\log A}{\eta_{\max}(T+1)}
+
30\eta_{\max}H^2
\right\}.
\]
\end{proof}

\subsection{Proof of the information-theoretic lower bound}
\label{app:information-theoretic-lower-bound}

We use the family defined in
Section~\ref{subsec:common-mdp-family}, with $K=S-1$ and the
uniform initial distribution specified in
Theorem~\ref{thm:common-information-lower-bound}.
Fix an agent that uses at most $N$ transition queries almost
surely and returns a Markov policy $\pi$.

For a fixed admissible parameter sequence, let $\mathbb P_\theta$
denote the distribution of the complete transcript under $M_\theta$,
including the agent's internal randomness, queries, oracle
observations, and returned policy. Write $\mathbb E_\theta$ for
the corresponding expectation. Let $C_{i'}^h(a_k)$ denote the
number of queries at $(h,s_{i'},a_k)$, and define
\[
C_{i'}^h:=\sum_{k=1}^{A}C_{i'}^h(a_k).
\]
Since queries at the absorbing state are excluded from these counts,
\[
\sum_{h=1}^{H-1}\sum_{i'=1}^{K}C_{i'}^h\le N
\qquad\text{almost surely}.
\]

\begin{lem}[Adaptive divergence decomposition]
\label{lem:adaptive-divergence-decomposition}
Let $M_\theta$ and $M_{\theta'}$ be two instances with the same
parameter sequence $(p^h,\alpha^h)_{h\in[H-1]}$. Then
\[
D_{\mathrm{KL}}(\mathbb P_\theta\|\mathbb P_{\theta'})
=
\sum_{h=1}^{H-1}
\sum_{i'=1}^{K}
\sum_{k=1}^{A}
\mathbb E_\theta[C_{i'}^h(a_k)]
D_{\mathrm{KL}}\!\left(
    B_{p_{\theta,i',a_k}^h}
    \,\middle\|\,
    B_{p_{\theta',i',a_k}^h}
\right).
\]
\end{lem}

\begin{proof}
Let $\tau\le N$ denote the number of queries. For $\ell\le\tau$,
write $X_\ell=(h_\ell,s_\ell,a_\ell)$ for the $\ell$-th query
and $Y_\ell$ for its response. If the agent stops before making
$N$ queries, append a fixed placeholder to each remaining position
in the transcript. These placeholders require no oracle queries
and have the same deterministic value under both instances.

Conditional on the complete preceding history, the agent uses
the same rules for generating its internal randomness, selecting
queries, and stopping under both instances. Their contributions
to the chain rule for relative entropy are therefore zero.
The conditional response distributions for an actual query
$X_\ell$ are
\[
P_\theta^{h_\ell}(\cdot\mid s_\ell,a_\ell)
\quad\text{and}\quad
P_{\theta'}^{h_\ell}(\cdot\mid s_\ell,a_\ell).
\]
The returned policy is determined by the observations and the
agent's internal randomness, so it contributes no additional
relative entropy. Consequently,
\[
D_{\mathrm{KL}}(\mathbb P_\theta\|\mathbb P_{\theta'})
=
\mathbb E_\theta\!\left[
    \sum_{\ell=1}^{\tau}
    D_{\mathrm{KL}}\!\left(
        P_\theta^{h_\ell}(\cdot\mid s_\ell,a_\ell)
        \,\middle\|\,
        P_{\theta'}^{h_\ell}(\cdot\mid s_\ell,a_\ell)
    \right)
\right].
\]
The parameter restrictions ensure that the transition
distributions from decision states have common support and
finite relative entropy. Queries at $g$ contribute zero because
its transition distribution is identical under both instances.
Since
\[
C_{i'}^h(a_k)
=
\sum_{\ell=1}^{\tau}
\mathbf 1\!\left\{
    X_\ell=(h,s_{i'},a_k)
\right\},
\]
grouping the remaining terms by their queried triples gives the
claimed identity.
\end{proof}

\begin{lem}[Local action-selection bound]
\label{lem:local-testing-bound}
Fix $h\in[H-1]$, $i'\in[K]$, and all labels other than
$\theta_{i'}^h$. For $k\in\{0,\ldots,A\}$, let $\mathbb P_k$
and $\mathbb E_k$ denote the transcript distribution and
expectation when $\theta_{i'}^h=k$. Then
\[
\frac1A\sum_{k=1}^{A}
\mathbb E_k\!\left[1-\pi^h(a_k\mid s_{i'})\right]
\ge
1-\frac1A
-
\sqrt{
    \frac{\mathbb E_0[C_{i'}^h]}{2A}
    D_{\mathrm{KL}}\!\left(
        B_{p^h-\alpha^h}
        \,\middle\|\,
        B_{p^h+\alpha^h}
    \right)
}.
\]
\end{lem}

\begin{proof}
The instances indexed by $0$ and $k$ differ only in the
transition distribution at $(h,s_{i'},a_k)$.
Lemma~\ref{lem:adaptive-divergence-decomposition} gives
\[
D_{\mathrm{KL}}(\mathbb P_0\|\mathbb P_k)
=
\mathbb E_0[C_{i'}^h(a_k)]
D_{\mathrm{KL}}\!\left(
    B_{p^h-\alpha^h}
    \,\middle\|\,
    B_{p^h+\alpha^h}
\right).
\]
Since $\pi^h(a_k\mid s_{i'})$ is a measurable function of the
transcript taking values in $[0,1]$, Pinsker's inequality implies
\[
\mathbb E_k[\pi^h(a_k\mid s_{i'})]
\le
\mathbb E_0[\pi^h(a_k\mid s_{i'})]
+
\sqrt{
    \frac{\mathbb E_0[C_{i'}^h(a_k)]}{2}
    D_{\mathrm{KL}}\!\left(
        B_{p^h-\alpha^h}
        \,\middle\|\,
        B_{p^h+\alpha^h}
    \right)
}.
\]
Average over $k$ and use
\[
\sum_{k=1}^{A}
\mathbb E_0[\pi^h(a_k\mid s_{i'})]=1,
\qquad
\frac1A\sum_{k=1}^{A}
\sqrt{\mathbb E_0[C_{i'}^h(a_k)]}
\le
\sqrt{\frac{\mathbb E_0[C_{i'}^h]}{A}}.
\]
Rearranging proves the result.
\end{proof}

\begin{lem}[Aggregation under the query budget]
\label{lem:query-budget-aggregation}
Let $\mu_\Theta$ be the distribution on label arrays with
independent entries satisfying
\[
\mu_\Theta(\theta_{i'}^h=0)=\frac12,
\qquad
\mu_\Theta(\theta_{i'}^h=k)=\frac1{2A},
\quad k\in[A].
\]
Let $\mathbb E_{\mu_\Theta}$ include the label array, oracle
observations, and agent's internal randomness, with value
functions evaluated in the selected instance. Then
\[
\begin{aligned}
\mathbb E_{\mu_\Theta}\!\left[
    V^{\star,1}(\rho)-V^{\pi,1}(\rho)
\right]
\ge \frac38\Bigg[
&\left(1-\frac1A\right)
\sum_{h=1}^{H-1}\alpha^h(H-h)
\\
&-
\sqrt{
    \frac{N}{AK}
    \sum_{h=1}^{H-1}
    (\alpha^h)^2(H-h)^2
    D_{\mathrm{KL}}\!\left(
        B_{p^h-\alpha^h}
        \,\middle\|\,
        B_{p^h+\alpha^h}
    \right)
}
\Bigg].
\end{aligned}
\]
\end{lem}

\begin{proof}
Retain the action-selection error $\mathrm{w}_\pi^h(i')$ from
Lemma~\ref{lem:family-policy-loss}. Apply
Lemma~\ref{lem:local-testing-bound} conditional on the other
labels. Averaging over those labels and using the concavity
of the square root gives
\[
\mathbb E_{\mu_\Theta}[\mathrm{w}_\pi^h(i')]
\ge
\frac12\left[
    1-\frac1A
    -
    \sqrt{
        \frac{
            \mathbb E_{\mu_\Theta}
            [C_{i'}^h\mid\theta_{i'}^h=0]
        }{2A}
        D_{\mathrm{KL}}\!\left(
            B_{p^h-\alpha^h}
            \,\middle\|\,
            B_{p^h+\alpha^h}
        \right)
    }
\right].
\]
Because $C_{i'}^h\ge0$ and
$\mu_\Theta(\theta_{i'}^h=0)=1/2$,
\[
\mathbb E_{\mu_\Theta}[C_{i'}^h]
\ge
\frac12
\mathbb E_{\mu_\Theta}
[C_{i'}^h\mid\theta_{i'}^h=0].
\]
Therefore,
\[
\mathbb E_{\mu_\Theta}[\mathrm{w}_\pi^h(i')]
\ge
\frac12\left[
    1-\frac1A
    -
    \sqrt{
        \frac{\mathbb E_{\mu_\Theta}[C_{i'}^h]}{A}
        D_{\mathrm{KL}}\!\left(
            B_{p^h-\alpha^h}
            \,\middle\|\,
            B_{p^h+\alpha^h}
        \right)
    }
\right].
\]

Under the uniform initial distribution,
Lemma~\ref{lem:family-policy-loss} gives, for every instance
and every realized returned policy,
\[
V^{\star,1}(\rho)-V^{\pi,1}(\rho)
\ge
\frac{3}{4K}
\sum_{h=1}^{H-1}\sum_{i'=1}^{K}
\alpha^h(H-h)\mathrm{w}_\pi^h(i').
\]
Taking expectations and substituting the preceding bound yields
\[
\begin{aligned}
\mathbb E_{\mu_\Theta}\!\left[
    V^{\star,1}(\rho)-V^{\pi,1}(\rho)
\right]
\ge \frac38\Bigg[
&\left(1-\frac1A\right)
\sum_{h=1}^{H-1}\alpha^h(H-h)
\\
&-
\frac1{K\sqrt A}
\sum_{h=1}^{H-1}\sum_{i'=1}^{K}
\alpha^h(H-h)
\sqrt{
    \mathbb E_{\mu_\Theta}[C_{i'}^h]\,
    D_{\mathrm{KL}}\!\left(
        B_{p^h-\alpha^h}
        \,\middle\|\,
        B_{p^h+\alpha^h}
    \right)
}
\Bigg].
\end{aligned}
\]
By Cauchy--Schwarz and the query-budget constraint, the double
sum is at most
\[
\sqrt{
    NK
    \sum_{h=1}^{H-1}
    (\alpha^h)^2(H-h)^2
    D_{\mathrm{KL}}\!\left(
        B_{p^h-\alpha^h}
        \,\middle\|\,
        B_{p^h+\alpha^h}
    \right)
}.
\]
Substitution proves the claim.
\end{proof}

\begin{proof}[Proof of Theorem~\ref{thm:common-information-lower-bound}]
Choose, at every $h\in[H-1]$,
\[
p^h=p:=\frac1{4H},
\qquad
\alpha^h=\alpha
:=
\min\!\left\{
    \frac1{8H},
    \frac1{32}\sqrt{\frac{KA}{N}}
\right\}.
\]
These choices satisfy
\[
0<\alpha<p,
\qquad
p+\alpha\le\frac{3}{8H}<\frac1{2H}.
\]

For $u,v\in(0,1)$, the definition of $B_p$ gives
\[
D_{\mathrm{KL}}(B_u\|B_v)
=
u\log\frac uv
+
(1-u)\log\frac{1-u}{1-v}
\le
\frac{(u-v)^2}{v(1-v)},
\]
where the inequality follows from $\log x\le x-1$. Hence
\[
D_{\mathrm{KL}}(B_{p-\alpha}\|B_{p+\alpha})
\le
\frac{4\alpha^2}{(p+\alpha)(1-p-\alpha)}
\le
32H\alpha^2.
\]
The choice of $\alpha$ therefore implies
\[
\frac{N}{AK}
D_{\mathrm{KL}}(B_{p-\alpha}\|B_{p+\alpha})
\le
\frac{H}{32}.
\]
Moreover,
\[
\sum_{h=1}^{H-1}(H-h)=\frac{H(H-1)}2,
\]
and
\[
\sum_{h=1}^{H-1}(H-h)^2
=
\frac{H(H-1)(2H-1)}6
\le
\frac2H
\left(\sum_{h=1}^{H-1}(H-h)\right)^2.
\]
Consequently,
\[
\sqrt{
    \frac{N}{AK}
    \alpha^2
    D_{\mathrm{KL}}(B_{p-\alpha}\|B_{p+\alpha})
    \sum_{h=1}^{H-1}(H-h)^2
}
\le
\frac{\alpha}{4}
\sum_{h=1}^{H-1}(H-h).
\]

Applying Lemma~\ref{lem:query-budget-aggregation} and using
$A\ge2$, we obtain
\[
\begin{aligned}
\mathbb E_{\mu_\Theta}\!\left[
    V^{\star,1}(\rho)-V^{\pi,1}(\rho)
\right]
&\ge
\frac{3\alpha}{32}
\sum_{h=1}^{H-1}(H-h)
\\
&=
\frac{3\alpha H(H-1)}{64}
\\
&\ge
\frac{3\alpha H^2}{128}
\\
&\ge
\frac{3}{4096}
\min\!\left\{
    H,\,
    H^2\sqrt{\frac{KA}{N}}
\right\}.
\end{aligned}
\]
This expectation is an average over finitely many label arrays.
Thus, at least one array $\theta$ satisfies the same lower bound
with expectation $\mathbb E_\theta$.

Finally, $K=S-1\ge S/2$, so
\[
\min\!\left\{
    H,\,
    H^2\sqrt{\frac{KA}{N}}
\right\}
\ge
\frac1{\sqrt2}
\min\!\left\{
    H,\,
    H^2\sqrt{\frac{SA}{N}}
\right\}.
\]
The theorem follows with the universal constant $c=3/8192$.
\end{proof}
\end{document}